\documentclass[lettersize,journal]{IEEEtran}
\usepackage{amsmath,amsfonts}
\usepackage{array}
\usepackage{tikz}

\usepackage{textcomp}
\usepackage{stfloats}
\usepackage{url}
\usepackage{verbatim}
\usepackage{graphicx}
\usepackage{multirow}
\usepackage{booktabs}
\usepackage{blkarray}
\usepackage{subcaption}
\usepackage{soul}
\usepackage{color}
\usepackage{cite}
\usepackage[ruled,vlined]{algorithm2e}
\begin{document}

\title{Improving Operational Efficiency In EV Ridepooling Fleets By Predictive Exploitation of Idle Times}

\author{{Jesper C. Provoost, Andreas Kamilaris, Gyözö Gidófalvi, Geert J. Heijenk, and Luc J.J. Wismans\\}
\IEEEauthorblockA{\textit{Dept. of Computer Science, University of Twente, Enschede, The Netherlands}
}}

\markboth{Non-peer reviewed paper}%
{Shell \MakeLowercase{\textit{et al.}}: A Sample Article Using IEEEtran.cls for IEEE Journals}

\maketitle

\begin{abstract}
In ridepooling systems which are operated with an electric fleet, charging is an essential but complex decision-making process. Most contemporary electric vehicle (EV) taxi services require drivers to make egoistic decisions on where, when and how long to charge, leading to decentralized ad-hoc charging strategies. Moreover, knowledge about the current state of the mobility system is often lacking or simply not shared between vehicles, making it impossible to make a system-optimal decision. As a consequence, transport and resource efficiency are likely to be suboptimal, impacting the profitability of the service from the operator's perspective.

Most existing approaches to intelligent charging control work do not combine time, location and duration into a comprehensive control algorithm or are unsuitable for real-time operation in networks with thousands of road segments. We therefore present a real-time predictive charging method for ridepooling services with a single operator. This method, called Idle Time Exploitation (ITX), predicts the periods where vehicles are idle and exploits these periods to harvest energy. It relies on Graph Convolutional Networks (GCNs) and a linear assignment algorithm to devise an optimal pairing of vehicles and charging stations, in pursuance of maximizing the exploited idle time. Our approach works on large-scale graph representations of the road network and enables fine-grained decision-making. We evaluated our approach through extensive simulation studies on real-world datasets from New York City. As a reference, multiple baselines were devised with varying levels of complexity. The results demonstrate that ITX outperforms all baseline methods by at least 5\% (equivalent to \$70,000 for a 6,000 vehicle operation) per week in terms of a monetary reward function which was modeled to replicate the profitability of a real-world ridepooling system. Moreover, ITX can reduce delays by at least 4.68\% in comparison with baseline methods and generally increase passenger comfort by facilitating a better spread of customers across the fleet. Our results also demonstrate that ITX enables vehicles to harvest energy during the day, stabilizing battery levels and increasing resilience to unexpected surges in demand. Lastly, compared to the best-performing baseline strategy, peak loads are reduced by 17.39\% which benefits grid operators and paves the way for more sustainable use of the electrical grid.
\end{abstract}

\begin{IEEEkeywords}
Ridepooling, Intelligent Fleet Management, Electric Vehicles, Charging, Machine Learning, Deep Neural Networks, Operations Optimization, Assignment Problem, Idle time
\end{IEEEkeywords}

\section{Introduction}
In shared transportation systems, multiple persons or goods can travel inside the same vehicle. It can be regarded as an essential component of sustainable transportation, since it increases the operational efficiency while relieving the burden on infrastructure and resources. The implementation of shared transportation systems can therefore provide substantial benefits with regard to operational cost, traffic congestion and consequently environmental impact \cite{agatz2010}.

Ridesharing is a prominent example of shared transportation which concerns the movement of people. The original concept of ridesharing, which is synonymous to \emph{carpooling}, allows drivers to offer a free seat in their vehicles if the destination of the fellow traveler is the same or along the route of the driver. This concept is suitable for long-distance travel but does not provide enough flexibility to serve intra-urban high-frequency trips within short notice. Ridepooling, which has been facilitated by the worldwide increase in connectivity (e.g. through smartphone usage), provides more flexibility by fulfilling on-demand trip requests in real-time. Customers of such a service can request a trip through their smartphone (by providing a real-time GPS location), after which the trip requests are assigned to a vehicle \cite{amey2011}. In this research, we regard ridepooling as a form of Demand Responsive Transport (DRT) in which a fleet of mixed-capacity shareable taxi vehicles is centrally managed and dispatched. The taxi vehicles are operated by dedicated human drivers who are financially compensated for serving trip requests. Present-day examples of such services are UberPool and Lyft Line. In the future, the real-time and ad-hoc nature of ridepooling services provide opportunities for the deployment of autonomous vehicles in urban transportation. In this research, however, we determine that vehicles are operated by employed human drivers as this is the prevailing practice in contemporary taxi and ridepooling systems \cite{amey2011}.

Electrification is one of the essential approaches to improve efficiency and sustainability in transportation. In 2021, it is expected that more than one million electric vehicles (EVs) will be sold in Europe alone \cite{schmidt2021}. Through various projects, the European Union promotes the use of electric vehicles in the transport sector, which could help to achieve its target of reducing $\text{CO}_2$ and particulate emissions \cite{eu2020}. However, from an operator's perspective, there are some drawbacks to operating a fleet of EVs as part of a transportation service (e.g. a taxi or ridepooling service). As of today, most optimization work in this area has focused on the placement of the charging locations, in order to ensure that the distance between the idle vehicles and the charging infrastructure is minimized \cite{asamer2016, xylia2017, yang2017, tu2016}. Other related work concerns the intelligent distribution of power across vehicles that are present at a charging station, e.g. so-called `adaptive charging' \cite{lee2020, xydas2016}. Even though such methods already greatly improve the efficiency of the charging infrastructure, they do not explicitly optimize the operational aspects of the service (such as routing and dispatching), nor do they consider charging as an integral element of the operational decision-making.   

One of the most important implications of EV fleets for the operator is the relatively limited range of the vehicles: in 2020, the electric passenger car with the highest range (i.e. the Tesla Model S Long Range) could drive just over 600 kilometers on a full battery while the dominant internal combustion engine (ICE) car in New York City taxi traffic, the Toyota Camry, could reach over 1,000 kilometers on a single tank \cite{wallbox2020}. An even more important drawback of EV fleets is the significant time increase that is required to charge the vehicles. A typical EV charging session (discounting overnight charging) would last approximately 30 minutes, while it would take at most 10 minutes to refuel an ICE vehicle \cite{chargedfuture2020}. Together with the degradation of the battery cells, this suggests that that it is essential to optimize the timing and duration of charging sessions, and to ensure that these decisions integrate well into the main operational processes, such as dispatching and repositioning \cite{clairand2019}.

When compared to EVs for private use, the challenges are more substantial for EVs in commercial transportation systems, as their required load is much larger and there are more spatial and temporal constraints to take into account \cite{clairand2019}. This is especially the case for ridepooling systems, where multiple customers can use the vehicle at the same time and where the dispatch and repositioning processes are thus more complex and intertwined with charging \cite{jung2012}. To determine what is the best time, location and duration for the vehicles to charge, one therefore needs to take into account not only the current state of the vehicles and charging infrastructure, but also all customers that need to be served en-route as well as the anticipated demands. To illustrate, it might be beneficial for a vehicle with a higher State-of-Charge (SoC) to charge early if a shareable demand is expected in the near future, such that the vehicle will be able to serve a large number of anticipated customers concurrently without having to charge. In another scenario, it could be optimal for a charging session to be aborted if a new request has been made and no other vehicles are nearby. Continuous trade-offs need to be made between serving customers on time and anticipating future demands while ensuring that the vehicles have sufficient energy to carry out their operations. In this research, we aim to develop a comprehensive control method which can produce real-time predictive decisions on charging behaviour and can be integrated with other decision-making processes. We therefore present a real-time predictive charging method for ridepooling services. This method, called Idle Time Exploitation (ITX), attempts to exploit idle time periods to charge the fleet and thus harvest energy. To the best of our knowledge, there is no existing method available which enables both real-time and predictive optimization of charging behaviour (i.e. when, where and how long) on a real-world scale level. The contributions of this paper can therefore be summarized as:

\begin{itemize}
    \item Optimizing decisions regarding charging behaviour, which involves \textbf{when}, \textbf{where} and for \textbf{how long} the vehicles charge. To the best of our knowledge, there does not exist a method which integrally optimizes the three decision variables by exploiting predicted idle times, simplifying the optimization task into a problem which can be solved in polynomial time. The proposed ITX methodology is shown to reduce operational costs from an operator's perspective while also reducing delays and improving comfort levels of passengers.
    \item Providing proactive charging opportunities which facilitate energy harvesting during the day. The resulting energy reserves can be used to satisfy peak demands more smoothly. Additionally, the power levels drawn from the grid are shown to be more stable over time and space, alleviating the burden on electricity infrastructure which paves the way for more sustainable use of the electrical grid.
    \item Ease of integration with other operational decisions, i.e. dispatching and repositioning, by letting control processes occur in modular and sequential fashion. This ensures faster runtimes (at the expense of global optimality), but also flexibility and robustness as the ITX method can be used alongside existing dispatching and repositioning frameworks.
    \item Real-time (in this study a one-minute interval) and predictive decision-making capabilities at the scale level and granularity of a large real-world city, with 6,000 vehicles and 8,500 roads. Simulation and policy generation are performed on graph representations of a road network, which benefits transferability of the system.
\end{itemize}

The rest of this paper is organized as follows. Section \ref{sec:relatedwork} describes relevant related work for this paper, and further defines the research gap which we aim to bridge. Section \ref{sec:methodology} explains the proposed methodology and the main components of the system design. In addition, this section describes the datasets used for the models and experiments. Section \ref{sec:experimentalsetup} then lays out the parameter settings for the models and simulator, as well as the design of the experiments. Section \ref{sec:results} provides the experimental results and highlights the main findings of this work. The conclusions, as well as opportunities for future research, are presented in Section \ref{sec:conclusion}.

\section{Related work}
\label{sec:relatedwork}
In recent years, several optimization techniques for EV charging have been proposed. However, most of the literature on this topic has focused on privately owned vehicles, typically assuming that these vehicles are used at a low frequency and charged at home during the night \cite{richardson2013, mukherjee2015, yi2020}. Furthermore, works on shared transportation systems deal mainly with the transport aspects, without considering the charging as a part of the transport operation. However, there is a selection of works available which focus on the optimization of electrification in transportation systems. In \cite{clairand2019}, Clairand et al. discuss that these works ``covered a broad range of objectives for ETs and EBs [i.e. electric taxis and buses], such as [positioning] and sizing charging stations, cost minimization, load unbalance minimization, planning of charging stations, and so on.'' The authors also argue that ``many of these works used optimization methods to solve their problems''. 

We aim to identify and categorize a selection of works that are relevant to charging optimization, in order to explore potential areas of interest and research gaps which are specific to ridepooling scenarios. Within this conceptual framework, we differentiate related work by the primary \emph{stakeholder} that benefits from the optimization, the \emph{type of transport} that it applies to, the \emph{decision variables} of the optimization task, the \emph{main objective} and the \emph{optimization technique} that is utilized. Additionally, we determine whether the proposed methods are centralized (i.e. a single agent approach), data-driven, predictive, and whether they could potentially scale towards large real-world transportation systems. It must be noted that it is hard to make a fair comparison of performance between the listed works, since they are implemented under highly different circumstances and for a variety of scenarios. Consequently, we will mainly observe results and evaluation regarding computational efficiency to assess the scalability of the proposed approaches. The resulting overview of related work is shown in Table \ref{tab:litoverview}. As a general rule, we can classify works based on the type of decision variable that is optimized: in Section \ref{sec:optcharginginfra}, the works which concern \emph{optimization of charging infrastructure} (i.e. positioning, pricing or scheduling) are described while Section \ref{sec:optchargingstrategies} describes the works which concern the \emph{optimization of charging strategies} (from the fleet operator's perspective). In the latter section, we explore predictive scheduling, heuristic and learning-based approaches before further narrowing down on the research gap that our proposed method aims to bridge.

\begin{table*}[t]
\resizebox{\linewidth}{!}{%
\begin{tabular}{lllllllllll}
\multicolumn{1}{l|}{\textbf{Author}}  & Stakeholder               & Type of transport & Decision variables              & Main objective                          & Approach & Centralized & Data-driven & Real-time & Predictive & Scalable \\ \hline
\multicolumn{1}{l|}{Xylia et al. \cite{xylia2017}}     & Public transport operator & Bus               & Charging station placement      & Emissions and energy consumption   & MILP              & \checkmark         & \checkmark         & $\times$        & $\times$         & \checkmark      \\
\multicolumn{1}{l|}{Houbbadi et al. \cite{houbbadi2019}}  & Public transport operator & Bus               & Charging scheduling             & Minimizing cost and battery aging  & NLP          & \checkmark         & $\times$          & $\times$        & $\times$         & \checkmark      \\
\multicolumn{1}{l|}{Yang et al. \cite{yang2017}}      & Infrastructure providers  & Taxi              & Charging station placement      & Minimizing infrastructure cost     & ILP             & \checkmark         & \checkmark         & $\times$        & $\times$         & $\times$       \\
\multicolumn{1}{l|}{Lee et al. \cite{lee2020}}       & Charging system operator  & Private           & Adaptive charging control       & Charging time, load distribution   & Framework             & $\times$          & \checkmark         & \checkmark       & $\times$         & \checkmark      \\
\multicolumn{1}{l|}{Yi et al. \cite{yi2020}}        & Charging system operator  & Private           & Adaptive charging control       & Load flattening                    & QP             & $\times$          & \checkmark         & \checkmark       & $\times$         & $\times$       \\
\multicolumn{1}{l|}{Wang et al. \cite{wang2021}}      & Charging system operator  & Private           & Pricing and scheduling          & Maximizing charging station profit & RL           & $\times$          & \checkmark         & \checkmark       & \checkmark        & \checkmark      \\
\multicolumn{1}{l|}{Jin et al. \cite{jin2014}}       & Charging system operator  & Private           & Charging scheduling             & Minimizing cost and delays         & LP           & $\times$          & $\times$          & \checkmark       & $\times$         & $\times$       \\
\multicolumn{1}{l|}{Sundström et al. \cite{sundstrom2010}} & Vehicle fleet operator    & Private (fleet)   & Charging scheduling             & Minimizing cost, power balancing   & LP, QP        & \checkmark         & $\times$          & $\times$        & $\times$         & $\times$       \\
\multicolumn{1}{l|}{Nguyen et al. \cite{nguyen2014}}    & Vehicle fleet operator    & Private (fleet)   & Charging scheduling             & Minimizing peak load               & LP           & \checkmark         & $\times$          & $\times$        & $\times$         & \checkmark      \\
\multicolumn{1}{l|}{Sassi et al. \cite{sassi2017}} & Vehicle fleet operator      & Private (fleet)       & Charging scheduling & Minimize charging cost    & Heuristic         & \checkmark         & $\times$         & $\times$       & $\times$        & \checkmark       \\
\multicolumn{1}{l|}{Shao et al. \cite{shao2017}} & Vehicle fleet operator      & Private (fleet)       & Routing and charging & Minimize travel and charging cost    & Genetic alg.         & \checkmark         & $\times$         & $\times$       & $\times$        & \checkmark       \\
\multicolumn{1}{l|}{Tian et al. \cite{tian2016}}  & Urban taxi provider & Taxi               & Charging station selection             & Minimizing waiting times  & Framework          & $\times$         & \checkmark         & \checkmark        & \checkmark         & $\times$      \\
\multicolumn{1}{l|}{Qin et al. \cite{qin2011}}  & Urban taxi provider & Taxi               & Charging scheduling             & Minimizing waiting times  & LP          & $\times$         & $\times$          & $\times$        & $\times$         & $\times$      \\
\multicolumn{1}{l|}{Wang et al. \cite{wang2013}}  & Urban taxi provider & Taxi               & Dispatching and routing             & Maximize time before charging  & Heuristic          & \checkmark         & $\times$          & \checkmark        & $\times$         & \checkmark      \\
\multicolumn{1}{l|}{Zhu et al. \cite{zhu2014}}  & Urban taxi provider & Taxi               & Charging scheduling             & Minimizing charging times  & Heuristic          & \checkmark         & $\times$          & $\times$        & $\times$         & $\times$      \\
\multicolumn{1}{l|}{Asamer et al. \cite{asamer2016}}    & Urban taxi provider       & Taxi              & Charging station placement      & Maximizing demand coverage         & MILP              & \checkmark         & \checkmark         & $\times$        & $\times$         & \checkmark      \\
\multicolumn{1}{l|}{La Rocca et al. \cite{larocca2019}} & Urban taxi provider      & Taxi       & Routing and charging & Income and waiting times    & Heuristic         & \checkmark         & $\times$         & $\times$       & $\times$        & \checkmark       \\
\multicolumn{1}{l|}{Tu et al. \cite{tu2016}}        & Urban taxi provider       & Taxi              & Charging station placement      & Maximizing demand coverage         & Genetic alg.             & \checkmark         & \checkmark         & $\times$        & $\times$         & \checkmark      \\
\multicolumn{1}{l|}{Shi et al. \cite{shi2018}}       & Ridepooling provider      & Ridepooling       & Charging scheduling             & Maximize transport efficiency      & MINLP        & \checkmark         & $\times$          & $\times$        & $\times$         & $\times$       \\
\multicolumn{1}{l|}{Kang et al. \cite{kang2016}} & Ridepooling provider      & Ridepooling       & Complete system design & Maximize profitability and feasibility    & SQP         & \checkmark         & $\times$         & $\times$       & $\times$        & $\times$       \\
\multicolumn{1}{l|}{Zhang et al. \cite{zhang2016}}     & Ridepooling provider      & Ridepooling       & Repositioning routes                         & Minimize waiting times             & LP           & \checkmark         & \checkmark         & \checkmark       & \checkmark        & $\times$       \\
\multicolumn{1}{l|}{Iacobucci et al. \cite{iacobucci2019}} & Ridepooling provider      & Ridepooling       & Repositioning and charging & Minimize cost and waiting times    & MILP         & \checkmark         & \checkmark         & \checkmark       & \checkmark        & $\times$       \\
\multicolumn{1}{l|}{Pettit et al. \cite{pettit2019}} & Ridepooling provider      & Ridepooling       & Routing and charging & Minimize cost and emissions    & DRL         & $\times$         & \checkmark         & \checkmark       & \checkmark        & $\times$       \\
\multicolumn{1}{l|}{Shi et al. \cite{shi2019}} & Ridepooling provider      & Ridepooling       & Dispatching and charging & Minimize societal cost    & DRL         & -         & \checkmark         & \checkmark       & \checkmark        & \checkmark       \\
\multicolumn{1}{l|}{Kullman et al. \cite{kullman2020}} & Ridepooling provider      & Ridepooling       & Dispatching, repositioning and charging & Maximize expected profit    & DRL         & -         & \checkmark         & \checkmark       & \checkmark        & $\times$       \\
                                      &                           &                   &                                 &                                    &              &             &             &           &            &         
\end{tabular}}
\caption{Overview of literature related to EV charging optimization.}
\label{tab:litoverview}
\end{table*}

\subsection{Optimizing charging infrastructure}
\label{sec:optcharginginfra}
\subsubsection{Pricing and scheduling}
Some available literature considers the main objectives of the optimization task to be related to the efficiency of the EV charging stations. Hence, these works regard the charging system operator as the main stakeholder. For instance, in \cite{wang2021}, Wang et al. propose a reinforcement learning (RL) approach to optimize the pricing and scheduling at a charging station. The main objective of this work was to increase the profit at an individual charging station. Other works aim to minimize charging cost \cite{wang2021, sundstrom2010} or delay time (i.e. time until a vehicle in the queue may charge) \cite{jin2014, zhang2016} at a charging station. Most of these works only optimize for individual charging stations and therefore neglect the implications of having a transportation system where the charging policies at multiple charging stations are interconnected with operational processes. In addition, many articles are focused on the implications of EV charging on the electrical grid, i.e. optimizing the power balance and peak load \cite{sundstrom2010, nguyen2014, yi2020}. However, none of these works consider the operational characteristics of a transportation service as the central objective of their optimization problem.

\subsubsection{Positioning of infrastructure}
There is a considerable number of papers which involve optimization of electrification-related decisions in (shared) transportation systems. Some of these papers consider the locations of the charging stations as the decision variable for the optimization problem. For instance, Asamer et al. \cite{asamer2016} propose a decision support system for placing charging stations such that the charging demand of electric taxi vehicles is satisfied. Tu et al. \cite{tu2016} implement a similar method, but incorporate temporal constraints to determine the optimal locations of the charging stations. Moreover, Xylia et al. \cite{xylia2017} propose a similar approach but targeted at charging infrastructure for electric buses in Stockholm. Another approach was developed by Gidofalvi and Yang \cite{gidofalvi2020}, who focus on electrification of road segments instead of conventional charging stations. The authors propose a route-based electric road network optimization (RENO) methodology which is shown to provide considerable cost savings for (shared) freight transportation systems. In this work, however, we still consider charging infrastructure in the form of fixed facilities where vehicles can be plugged in, given that such systems are most often used in practice \cite{chargedfuture2020}.

\subsection{Optimizing charging strategies}
\label{sec:optchargingstrategies}
Other related work also considers electric buses, but aims to optimize the charging strategies from the fleet's perspective instead of the placement of charging infrastructures. In \cite{houbbadi2019}, Houbbadi et al. propose a nonlinear programming approach to optimize the charging strategy of an electric bus fleet. Using a numerical simulation, the authors demonstrate that their approach minimizes electricity cost and the battery aging. Even though their work does consider operational aspects of the bus service, the authors only optimize for overnight charging schedules. This scenario is considerably different from a taxi or ridepooling service where vehicles serve a larger variety of trips which do not follow a planned route. Given that battery capacities in ETs (i.e. electric taxis) are smaller and that longer overall distances are covered \cite{clairand2019}, they will need to charge in between trips, leading to additional constraints and objectives which should be taken into consideration.

There are other works that attempt to optimize EV charging from a transport-operational perspective. However, many of these works use fixed parameters such as the battery level to determine when to start and finish a charging session. For instance, Tian et al. \cite{tian2016} propose a charging station recommendation system for EV taxis. The times at which the vehicles charge are predetermined based on a real-world dataset of recharging events, and are therefore not part of the decision variables for optimization. The authors also assume that a vehicle must remain at the charger until the battery is completely full. However, since less power is supplied when the battery level approaches 100\%, charging sessions with varying duration can be more efficient for the overall transport operation. Similarly, other authors propose methodologies that do not consider the duration of charging as a decision variable \cite{qin2011, wang2013, zhu2014}. This presents further opportunities for the optimization of charging schedules within the operational decision-making process.

In \cite{shi2018}, Shi et al. propose a linear programming solution to determine the optimal routes for an EV fleet given a set of pickup and dropoff requests and the battery states of the vehicles. They incorporate the spatio-temporal relationship between vertices in the graph and consider minimization of the waiting time and the travel distance as the objective of their optimization problem. The authors demonstrate that the proposed method can find optimal routes for small-scale problems. However, they add that the optimization problem is NP-hard and ``does not scale well with the number of customers and EVs''. Therefore, this approach is likely inadequate for real-time optimization of charging strategies with the scale level and granularity that are required by full-scale ridepooling services. Additionally, the authors do not consider the possibility of letting vehicles charge when they are idle, even though this is arguably the most advantageous moment to charge from an operational as well as a customer's perspective.

\subsubsection{Predictive scheduling}
In contrast to this approach, Zhang et al. \cite{zhang2016} propose an optimization method which attempts to predictively control repositioning strategies, with charging being a constraint. Hence, the scheduling of charging sessions was not optimized and vehicles would charge at maximum power as soon as they connect to charging stations. In \cite{iacobucci2019}, Iacobucci et al. therefore extend the former method by integrating the scheduling of charging as a decision variable in the optimization problem.  A numerical simulation performed with real-world data from Tokyo suggests that their approach allows ``efficient optimization of both aspects of system operation''. Yet, the authors mention that their approach is limited in scale, as the computational complexity grows more than linearly with the simulation size. As a result, solving the optimization problem becomes ``infeasible for more than a few tens of vehicles'', which makes the approach unsuitable for real-time optimization. The authors suggest that heuristics-based optimization models can be regarded a scalable alternative for large scenarios. Another limitation of their work is that charging stations are modelled in a relatively simplistic fashion. For instance, the authors assume that there are no congestion constraints at charging stations, while in reality most stations will be able to serve only a few vehicles simultaneously. Moreover, the authors consider a simple linear battery model and allow vehicles to charge at any vertex in the network, which also does not represent a realistic scenario.

\subsubsection{Heuristic approaches}
Similarly, other articles that focus on the optimization of charging scheduling often employ mathematical optimization techniques such as the ones previously mentioned \cite{kang2016, sassi2017}. In practice, this usually means that such methods are either too computationally complex to ensure scalability or neglect essential constraints. An alternative is to use heuristic-based methods \cite{sassi2017, shao2017, larocca2019}, which produce fast and feasible solutions but are unable to find a globally optimal solution. Furthermore, heuristic approaches are relatively inflexible and therefore can lose their effectiveness when operational processes or priorities change. For instance, a heuristic which mainly aims at minimizing customer waiting times could lose its value when an operator decides that it is more important to focus on maximizing fleet utilization. Learning-based approaches could be regarded as a suitable alternative due to their ability to adapt to highly complex environments while providing full flexiblity in defining the objective function.

\subsubsection{Learning-based approaches}
In \cite{pettit2019}, Pettit et al. propose a learning-based approach in the form of deep reinforcement learning (DRL). The authors attempt to optimize the driving and charging policy for an agent, i.e. a ridepooling EV. Simulations on real-world data suggest that their agent outperforms the heuristic baselines. However, the authors only consider a single vehicle in the environment and focus mostly on the short-term reward from charging. Hence, the interactions between vehicles as well as the system-wide operational influences are neglected in this research. Therefore, the approach is arguably unsuitable for transportation systems with a central operator, where it is more essential to optimize towards the system-wide benefits. Shi et al. \cite{shi2019} also propose a reinforcement learning approach, but their work employs a multi-agent framework where the interactions between vehicles in the fleet are simulated. The framework produces separate actions for every vehicle which are then used by a centralized agent to make decisions on dispatching and recharging. The authors demonstrate that their approach outperforms benchmark algorithms in terms of operational costs and customer delays. However, Kullman et al. \cite{kullman2020} argue that a limitation of this work is that no repositioning is taken into account in the decision-making process. Therefore, they propose a multi-agent DRL method which simultaneously makes repositioning and dispatching decisions. Even though the authors demonstrate that their method outperforms the baseline approach, they note that the performance decreases as the number of vehicles and requests grows. This suggests that their approach does not scale well with large fleets and complex environments. Additionally, the authors do not model the environment in high detail: they divide Manhattan, New York City into 61 taxi zones. In practice, demands and traffic flows might differentiate significantly between roads and crossings, and hence a graph representation of the environment will yield more accurate simulations, and consequently, a more fine-grained policy. Besides, using a graph representation would be beneficial for the transferability of the proposed solution, as taxi areas are defined differently per city whereas graph representation can be composed for every road network from ubiquitous sources such as OpenStreetMap. Another limitation of the work by Kullman et al. is that repositioning and charging are regarded as a single action. This prevents vehicles from being able to reposition to or via any other location than the 302 charging stations. It also neglects the possibility for vehicles to follow a maximum-likelihood repositioning path where new customers can be picked up along the way. Furthermore, the duration of a charging session is not directly determined by the DRL agent, even though we argue that it potentially has a large impact on the overall operational efficiency. Overall, this suggests a need for a system which can produce real-time and predictive decisions on charging behaviour and integrate the three decision variables \emph{when}, \emph{where} and \emph{for how long} into a comprehensive control method. It also emerges that the influence of charging times, locations and duration on the operational performance of a ridepooling system is not yet understood. We have thus identified a research gap on the integration of charging optimization into the operational decision-making processes in order to maximize operational efficiency and hence the profitability for the operator. Our proposed Idle Time Exploitation (ITX) method aims to bridge this research gap by simplifying the optimization problem, intuitively making use of idle times as well as ensuring transferability in real-world scenarios with thousands of vehicles moving on a fine-grained road network.

\section{Methodology}
\label{sec:methodology}
\subsection{Overview}
The methodology can be divided in five different components. First, we preprocess and clean the real-world trip datasets and map the trips to the vertices in the network under study. Subsequently, we describe the construction of an environment which simulates the movement of ridepooling vehicles on the selected road network, and explain the algorithms that are used to optimally dispatch the vehicles to the requests and reposition them when they are idle. We then propose the algorithm which is used for optimizing the charging schedules of the ridepooling vehicles. In order to achieve this, we first describe the Graph Convolutional Network model that is trained to predict idle times, upon which the charging optimization algorithm acts.

\subsection{Datasets}
\label{sec:datasets}
The simulations and control mechanisms are based on a real-world taxi trip request dataset which contains more than 100 million taxi trip requests from the city of New York City, USA. The data is released by the New York City Taxi and Limousine Commission (NYC TLC) on a monthly basis \cite{tlctaxidata}. We decide to use `traditional' taxi data instead of specific data about ridepooling for a variety of reasons. First of all, the volume of data is much larger for regular taxi systems, which benefits the accuracy of our simulations and also facilitates assessment of the proposed approach and its scalability in a real-world scenario. Moreover, the high usage of TLC taxi data in related work means that results can be verified and compared with other approaches. We argue that data of regular taxi trips can be used to represent realistic demands. This approach provides advantages for the majority of real-world scenarios where trip data about ridepooling is not (yet) available. However, it must be noted that the volume of trip requests is likely not a completely accurate method to measure demand, due to the limitations in availability of New York City taxicabs. As a result, trip request data will likely produce lower demand estimates than there would be in reality. While it is difficult to provide more accurate estimates of demand, one should take into account that results could be influenced marginally, e.g. delays at peak times being slightly lower than they would be in reality.

The graph representing the road network was retrieved from OpenStreetMap using the OSMnx Python library \cite{boeing2017}. Upon initialization of the graph, it was saved as a GraphML file in order to be further processed and utilized with the Igraph Python library \cite{igraph}.

\subsection{Preprocessing}
The raw trip data from TLC comprises a considerable volume, i.e. more than 10 GB for the complete year-long period. Therefore, it is an essential task to clean and process the data to decrease both the time and space complexity of the simulated environment. Since mid-2016, TLC uses area codes instead of coordinates to describe the origin and destination of trips. This makes it harder to accurately match trips to vertices in the graph. Hence, we use one year of trip data from July 2015 until July 2016, where the coordinates are still available. The relevant attributes that are available for every trip are: \emph{pickup date/time}, \emph{pickup coordinates}, \emph{dropoff coordinates} and \emph{number of passengers}. 

Only the Manhattan area of New York City was selected in order to have a clear and well-defined study-area. Subsequently, irregularities and redundancies were treated: duplicate vertices and edges were consolidated and small isolated sections were removed. Dead-ends were removed from the graph in order to prevent vehicles in the simulator environment from getting stuck. The resulting directed graph contains 3,555 vertices and 8,535 edges.

Based on the road network that is represented by the graph, only the trips within Manhattan (i.e. both the pickup and dropoff coordinates located in Manhattan) were queried from the trip request dataset. Also, trips with an average speed of lower than 1 km/h or more than 100 km/h are considered to be invalid and removed from the dataset. The trips are matched to the vertices of the graph in accordance with the methodology proposed in Section \ref{sec:tripmatching}.

\subsubsection{Matching trips to vertices}
\label{sec:tripmatching}
We start by defining the set $R$ of trip requests. An individual trip request $r \in R$ contains multiple attributes: a pickup timestamp, pickup coordinates and dropoff coordinates. We also define the road network as a directed graph $G$, with a set $N$ of vertices and a set $E$ of edges. We simply match the trip's pickup coordinates with the nearest vertex based on the haversine distance $d(P_1, P_2)$, and do the same for the dropoff coordinates. Upon implementation, the performance of this algorithm is enhanced using vectorization in Python.

The algorithm yields an origin-destination mapping $(n_o, n_d)$ for all $r \in R$, with $n_o$ being the pickup vertex and $n_d$ being the dropoff vertex. With the resulting origin-destination mapping, we then perform the following tasks:

\begin{itemize}
    \item We determine the travel times (i.e. the time difference in seconds between pickup and dropoff time) for all origin-destination pairs of vertices in the dataset. Accordingly, we train a time and space-dependent XGBoost model which can predict the travel times between an arbitrary pair of vertices in $G$. This model uses the origin and destination coordinates, euclidean distance between the two points, time of day and day of week as inputs. After training the model with similar setings as used in \cite{kaggle}, the $R^2$ score on the test set is $0.81$. At fixed intervals (hourly in our simulator), travel times are predicted for all combinations of vertices in the road network, i.e. every combination of coordinate pairs.
    \item We save the original-destination mapping to a dataset, which will be replayed in the simulator. This way, the trip request data can directly be coupled to the network vertices in the simulator such that vehicles can be assigned to the requests, and can move along the origin and destination vertices, respectively. The related processes are further explained in $(1)$, $(2)$, $(3)$ and $(6)$ of Section \ref{sec:simulation}.
\end{itemize}

\subsection{Simulation}
\label{sec:simulation}
\begin{figure*}[t]
  \centering
  \includegraphics[width=0.84\linewidth]{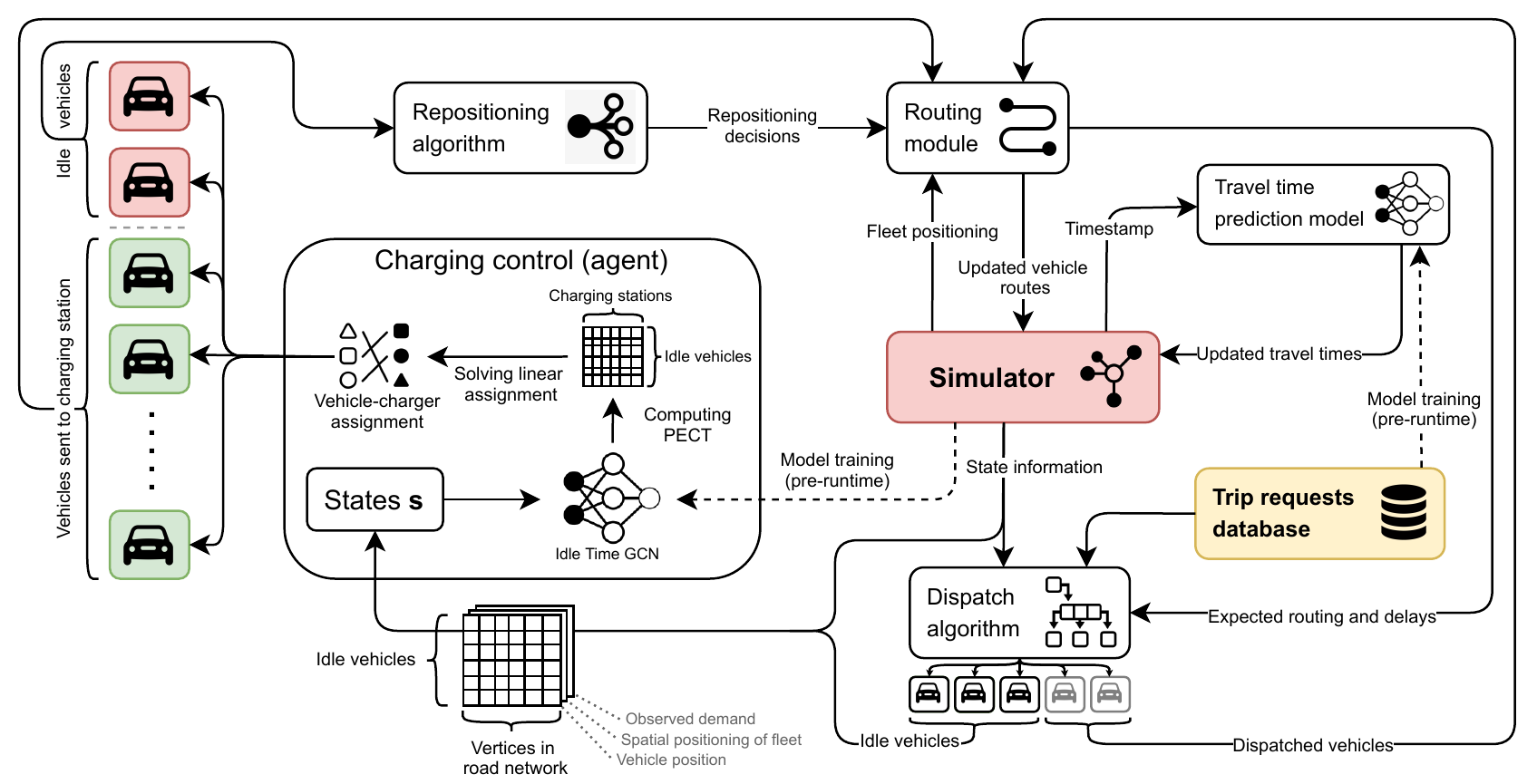}
  \caption{Schematic diagram of simulator processes and modules}
  \label{fig:systemdiagram}
\end{figure*}

We have developed a simulator which can simulate passenger demand and the trajectories of a ridepooling fleet at intervals of 1 minute. All trip requests in the simulation are exactly reproduced according to the trip request dataset as described in Section \ref{sec:datasets}. Based on the planned paths (i.e. either when serving requests or repositioning), vehicles move from vertex to vertex inside the simulated road network. The path of a vehicle is defined as a queue containing a sequence of vertices, where the first vertex in the queue is removed once that vertex has been reached. Their speed of movement along the edges is determined by the predicted travel times, which are recomputed at an hourly interval. Vehicles are objects which individually keep track of the current vertex and the time that they have spent at that vertex - if the vehicle has been there longer than the travel time towards the next vertex in the queue, the vehicle will move to the next location and the first vertex will be removed from the queue. This process of moving vehicles through the network based on travel time estimations forms the fundamental core of the simulator. In Figure \ref{fig:systemdiagram}, a comprehensive diagram of the simulator processes and modules is shown.

\subsubsection{Assumptions and definitions}
Several assumptions were made upon modeling the ridepooling system and the corresponding simulator. These assumptions are necessary to ensure that the study is clearly defined and bounded. Also, they help to maintain acceptable complexity levels, as well as consistency and stability of the experiments.

\begin{itemize}
    \item All customers of the ridepooling service are willing to share their trip with other customers. That is, every customer can be theoretically combined with any other, subject to the maximum capacity and current path of the vehicle.
    \item Requests which cannot be served within 5 minutes are neglected. This means that, with a one-minute a simulation interval, there are five opportunities for a vehicle to be assigned to a request by the dispatch algorithm.
    \item Demand is assumed to be equal to the actual observed taxi trips in the dataset, as previously explained in Section \ref{sec:datasets}.
    \item The traffic status in the simulator (e.g. many vehicles moving along an edge) does not influence the travel time predictions, hence avoiding the occurrence of feedback loops. Moreover, movement of other (non-fleet) traffic is not simulated.
    \item The initial positioning of the fleet is determined randomly upon initialization using a uniform distribution. Hence, every vehicle will be placed at a random vertex in $G$, with all vertices having the same likelihood of being selected.
    \item The locations of the charging stations are fixed, and determined in accordance with Section \ref{sec:chargermodeling}. Also, the assumption is made that charging infrastructure is only utilized by the fleet of the ridepooling service. The presence of other (e.g. consumer) vehicles is therefore not taken into account.
    \item If a vehicle runs out of battery, it will hold position for 60 minutes, after which it will be towed to the nearest charging station. Further detail about towing costs is provided in Section \ref{sec:vehiclemodeling}.
\end{itemize}

\subsubsection{Processes}
\label{sec:simulationprocesses}
During a single iteration in the simulator, which equals a one-minute time interval, several actions are performed:

\begin{enumerate}
    \item \textbf{Request retrieval} Retrieve the requests that occur at the current timestep from the dataset, merge with requests that could not be served during the last five minutes. Requests which cannot be served after 5 minutes are neglected, leading to a lower trip acceptance rate.
    \item \textbf{Request handling} Handle the outstanding requests and compute the optimal vehicles to serve them, based on the number of vacant seats, distance from the new customer and the expected delay for existing customers. If the request has been pending for more than 5 minutes, it is rejected.
    \item \textbf{Vehicle dispatching} Dispatch the vehicles which have accepted a trip request in Step 2. Compute their updated path, number of remaining vacant seats and estimated travel time.
    \item \textbf{Charging control} This is the primary action under study in this work. First, the idle vehicles (i.e. with no occupied seats and no planned route) are selected. Subsequently, the charging control algorithm (proposed in Section \ref{sec:chargingcontrol}) determines which idle vehicles should charge, and to which charging stations they will be sent. Moreover, when a vehicle is out of battery, it will need to hold position for 60 minutes, after which it will be towed to the nearest charging station.
    \item \textbf{Vehicle repositioning} After the charging control algorithm determines which vehicles are sent to a charging station, the remaining idle vehicles (i.e. with no occupied seats and no planned route) are repositioned along a route which is computed by the repositioning algorithm. Repositioning is based on the current position of a vehicle, the expected demand and positioning of other vehicles.
    \item \textbf{Moving vehicles} Based on the sum of time spent at the current vertex and the time advancement in the simulator, it is determined which vehicles should move to their next vertex. If a vehicle moves to a new vertex, the corresponding amount of energy is drawn from the battery. Subsequently, we check if the vehicle has reached one of the following:
    \begin{enumerate}
        \item The \emph{pickup vertex} of its customer, in which case the customer is picked up and the number of occupied seats is increased.
        \item The \emph{dropoff vertex} of its customer, in which case the customer is dropped off and the number of occupied seats is decreased. If the vehicle is then empty, it becomes \emph{idle} and will be dispatched or repositioned in the next iteration.
    \end{enumerate}
    
    \item \textbf{Saving metrics} The relevant metrics are stored in memory. All measurements are saved to disk periodically at an interval of 60 minutes (simulator time).
\end{enumerate}

\subsubsection{Vehicle modeling}
\label{sec:vehiclemodeling}
\begin{figure}[h]
  \centering
  \includegraphics[width=0.92\linewidth]{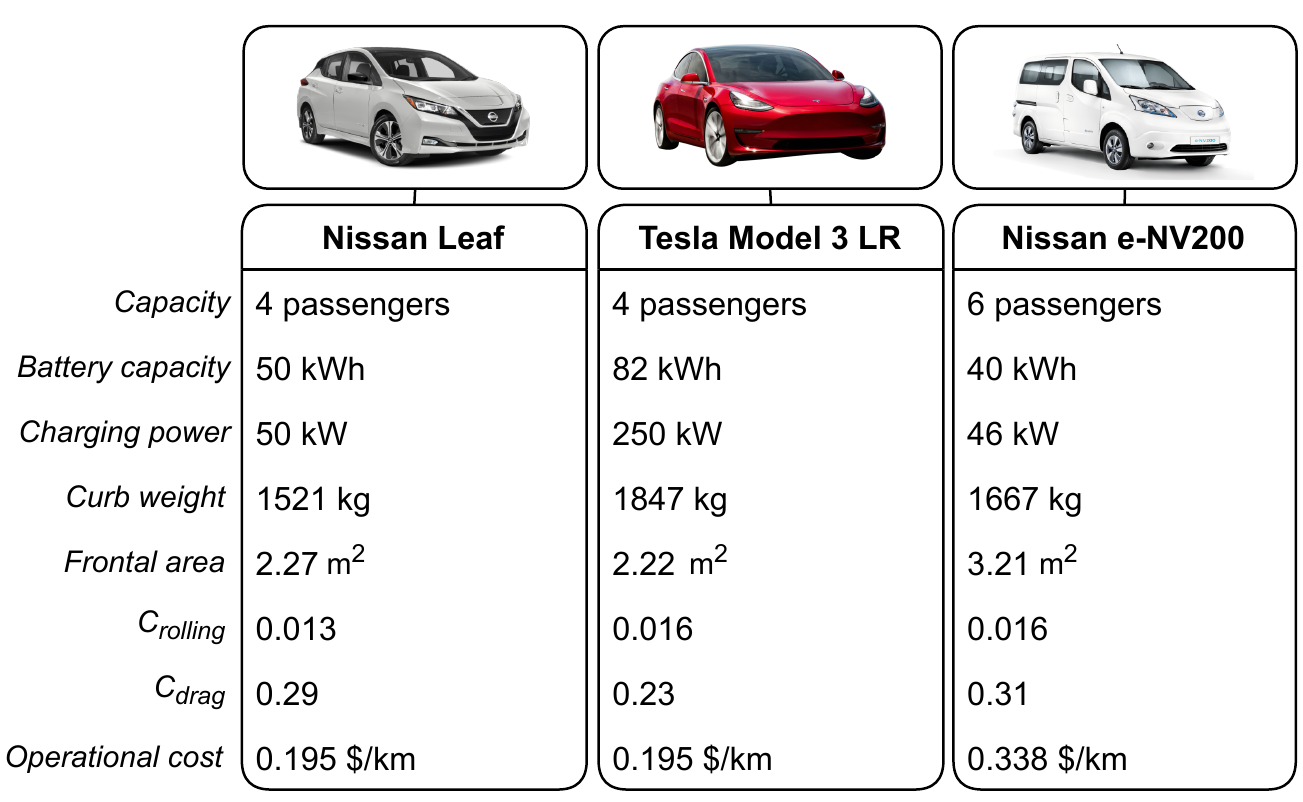}
  
\caption{Specifications of parameters for every EV model}
\label{fig:carspecs}
\end{figure}

To facilitate the assessment of our method's robustness, we attempt to model a heterogeneous fleet with a variety of vehicle types. The main objective is to bring variance in battery capacity, charging power and seating capacity of the fleet. Therefore, we decided to model a compact hatchback (Nissan Leaf), a sedan (Tesla Model 3 LR) and a van (Nissan e-NV200) using real-world data to build energy consumption and operational cost models that are reasonably accurate. The parameters for each vehicle model \cite{nissanleafspecs, teslamodel3specs, nissanenv200specs}, which are used in the energy consumption step (6) of Section \ref{sec:simulationprocesses} and the reward function of Section \ref{sec:rewardfunction}, are displayed in Figure \ref{fig:carspecs}. The simulator environment allows for the initialization of any number of vehicles of each vehicle type.

The energy consumption behaviour of the battery is determined by the curb weight, number of passengers, frontal area and friction coefficients of the respective vehicle type, as described by the following equations \cite{kirkpatrick}:
    
\begin{gather}
    P = \frac{1}{2} \cdot 1.225 \cdot C_{d} \cdot A \cdot v^3 + 9.81 \cdot C_r \cdot m \cdot v \quad [W] \\
    E = \frac{P \cdot t}{3.6 \cdot 10^6} \quad [\text{kWh}]
\end{gather}

In this equation, $C_{d}$ denotes the drag coefficient, $C_{r}$ the rolling resistance coefficient and $A$ the frontal surface, all of which are dependent on the vehicle type (see Section \ref{sec:vehiclemodeling} for the specifications per type). For every vehicle, we set the idle power $P_{idle}$ to 1500 W. The mass $m$ is determined by the sum of the vehicle's curb weight and the number of passengers multiplied by 80 kg. The travel time $t$ in seconds and speed $v$ in m/s are derived from the predicted travel time and the distance of the edge that was traversed.

The charging behaviour of the battery is determined by the battery capacity and the specified charging power $P_{max}$. In a real-world scenario, the charging rate (i.e. the supplied power) tapers off after reaching a State-of-Charge (SoC) of approximately 70\% \cite{kullman2020}. Therefore we model that, between a SoC of 70\% and 100\%, the supply power decreases linearly from the vehicle's maximum charging power to zero. 

\begin{equation}
    P_{charge}(SoC)= 
    \begin{cases}
        P_{max},& \text{if } SoC\leq 0.7\\
        \frac{1}{0.3}P_{max} - \frac{1}{0.3}P_{max} \cdot SoC,              & \text{otherwise}
    \end{cases}
\end{equation}

The charging curve therefore is asymptotic and it will be impossible to fully charge the battery. Therefore, independent of the requested charging time, the vehicle will be uncoupled when the SoC is equal to or larger than 99\%.

\subsubsection{Charger modeling}
\label{sec:chargermodeling}
\begin{figure}[h]
  \centering
  \includegraphics[width=0.77\linewidth]{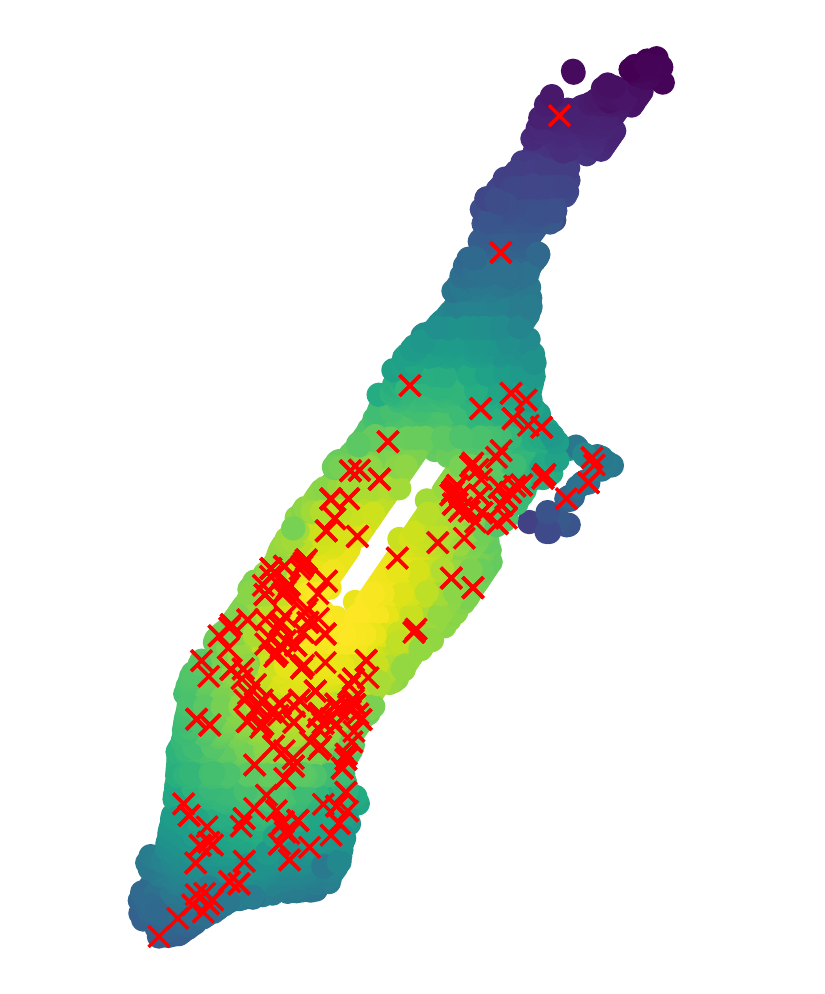}
  
\caption{Random allocation of 500 chargers in Manhattan, with underlying heatmap of closeness centrality}
\label{fig:evselocations}
\end{figure}
The locations of chargers in the simulator are determined using a probabilistic procedure. First, since EV chargers are often positioned on parking areas, we download the locations of all off-street parking areas in Manhattan, New York City using OpenStreetMap. The resulting coordinates are then matched to the vertices of the road network in accordance with Section \ref{sec:tripmatching}. Subsequently, we compute the closeness centrality using the Igraph library \cite{igraph}. Our reasoning for this is that more chargers are placed in denser and closely connected areas of the city since there will be a higher energy demand due to increased vehicle movements through these areas. For all vertices at which a parking area is located, the closeness centrality score is added to an array. The values in this array are then normalized and used as probabilities to perform random sampling of the charger locations. The sampling is repeated $k$ times (depending on how many chargers should be placed, which is a simulator parameter) with replacement. An example of $k = 500$ randomly generated chargers according to this procedure is visible in Figure \ref{fig:evselocations}. Please note that we define a vertex with one or more chargers as a \textbf{charging station}, i.e. a charging station is a location where vehicles can plug in to one or more chargers. 

The charger model is based on the Tesla Urban Supercharger network, which is currently in operation in Manhattan and consists of 72 kW DC fast chargers \cite{teslaurbansupercharger}. During a charging session, the supply power is therefore determined by the vehicle battery but limited at 72 kW. The charger is able to deliver its full power continuously, i.e. independent of other vehicles being charged at the same charging station. However, depending on whether the SoC of the battery is above or below the tapering point, the rate at which the battery is charged might decrease (see Section \ref{sec:vehiclemodeling}). We implement a queuing mechanism at the charging stations: when the control algorithms decide that a vehicle should charge, it will store internally the ID of the charging station as well as the planned charging time. Once the vehicle arrives at the charging station, it will be placed in a queue. Once a charger within the charging station becomes available, it is assigned to the next vehicle for the requested time period (or until the battery is 99\% full). The charging stations therefore operate according to a \emph{first in, first out} (FIFO) mechanism. Even though this might result in scenarios where a vehicle can charge later than initially expected upon decision-making, we argue that such a system represents the real-world better, since most contemporary charging infrastructure operates according to a first come, first serve principle. To mitigate the issues, we ensure that all vehicles have access to the expected waiting times at the charging stations, based on queue lengths as well as all vehicles that are currently underway to the charging station.

\subsubsection{Reward function}
\label{sec:rewardfunction}
The main reward function is modeled to represent the operating profit of the ridepooling service. Since a ridepooling service on the scale level of this research does not exist today, there is limited information on the exact cost items that such a service would face in a real-world scenario. Therefore, we base our monetary reward function on the existing regular taxi services in New York City, as well as ridehailing services such as Uber. We determine that the operating cost consists of multiple components: the \textbf{vehicle operational cost} $C_{op}$ (in \$/km), the \textbf{charging cost} $C_{charge}$ (in \$/kWh) and the \textbf{towing costs} $C_{tow}$ (in case a vehicle gets stranded). The vehicle operational cost is defined per vehicle type in Section \ref{sec:vehiclemodeling} and based on the work of Bösch et al. \cite{bosch2018}. The charging cost is set to $\$0.40$ per kWh \cite{electrekchargers, revelchargers}. The base towing cost is $\$125$ plus $\$2.50$ per kilometer to the nearest charging station \cite{towingcost}. Since only ongoing costs (i.e. those that are influenced by resource usage and energy consumption) are relevant in this research, we disregard other costs which are hard to quantify within our experiments, such as write-off costs, the purchasing cost of the fleet or the construction cost of the charging infrastructure.

The revenue of the ridepooling service is generated by succesfully serving requests. Upon acceptance of a new ride request, the fare of the ride is computed. The fare formula is based on the UberX fare systems in New York City \cite{uberxpricing}, where $F_r$ is the total fare in US dollars of request $r$, $t_{travel, r}$ is the estimated direct travel time in minutes between the pickup and dropoff vertices of $r$ and $d_{travel, r}$ is the network distance in kilometers. The minimum fare is $\$7.00$.

\begin{equation}
    F_r = max\{2.55 + 0.35 \cdot t_{travel, r} + 1.09 \cdot d_{travel, r}, 7\} \quad [\$]
\end{equation}

Once a customer has been dropped off at the planned destination within a certain delay (a parameter referred to as the \emph{maximum delay}), the fare $F$ is allocated. However, 75\% of the fare is paid out to compensate the driver, leaving 25\% (based on Uber's service fees \cite{uberservicefee}) for the operator of the service. Hence, 25\% of the cumulative fare is added as a component of the reward function, as this is the actual amount that is earned by the ridepooling operator. The total reward $R$ in minute $t$, with $R^{*} \subseteq R$ being the subset of requests which were served within the maximum delay and $V$ being the set of vehicles in the fleet, can be defined as follows:

\begin{equation}
    R_t = 0.25 \sum_{r \in R^{*}} F_{r} - \sum_{v \in V} (C_{op, v} + C_{charge, v} + C_{tow, v}) \quad [\$]
\end{equation}

\subsection{Dispatch and repositioning control}
This paper makes use of a sequential decision-making approach, i.e. separating the charging decision-making process from the dispatch and repositioning processes. We realize that this introduces potential suboptimality. However, we argue that this approach provides advantages in the form of modularity (providing more flexibility and transferability for operators) as well as performance, since the optimization problems are often NP-hard and simplification of the problems is essential to maintain acceptable computation times. In this section, we describe the dispatch and repositioning control algorithms that were used in the simulations.

\subsubsection{Request handling and dispatch}
Before a vehicle can be dispatched to a customer, the incoming requests should be handled and assigned to the fleet, such that the combined delays for all customers are minimized. This forms a complex optimization problem which becomes intractable when the fleet consists of thousands of vehicles serving customers in a large-scale road network. Therefore, one should find a sufficiently good, but potentially suboptimal solution using (meta)heuristic algorithms.

For the handling of requests and dispatch, which is not the main focus of this research, we employ the \emph{Dynamic Trip-Vehicle Assignment} method which was proposed by Alonso-Mora et al. \cite{alonsomora2017}. The authors devise a dispatching method which is suitable for large-scale ridepooling system. It allows large groups of riders to be matched to a fleet of shared vehicles in real-time. The proposed method consists of multiple steps leading to an integer linear optimization which provides an anytime optimal assignment. First, a pairwise request-vehicle (RV) graph is formed, consisting of both vehicles and requests. Two requests $r_1$ and $r_2$ are connected in the graph if they can potentially be combined. This is the case when a virtual (unoccupied) vehicle could pick up and drop off both requests with both customers incurring a delay of less than a certain constraint. We set this constraint to 5 minutes, matching the definition of what we consider to be `on-time' (an identical definition is used in subsequent parts of this research). If the combination of $r_1$ and $r_2$ is feasible, a cost is assigned as weight to the edge $e(r_1, r_2)$. In this work, we define this cost to be the minimum total delay that is incurred by both customers. This minimum delay is computed simply by an exhaustive search, which can be solved quickly since there are only six possible pick-up/drop-off sequences for two requests:

\[(1) \quad v_{r_1}^{PU} \rightarrow v_{r_1}^{DO} \rightarrow v_{r_2}^{PU} \rightarrow v_{r_2}^{DO}\]
\[(2) \quad v_{r_2}^{PU} \rightarrow v_{r_2}^{DO} \rightarrow v_{r_1}^{PU} \rightarrow v_{r_1}^{DO}\]
\[(3) \quad v_{r_1}^{PU} \rightarrow v_{r_2}^{PU} \rightarrow v_{r_1}^{DO} \rightarrow v_{r_2}^{DO}\]
\[(4) \quad v_{r_1}^{PU} \rightarrow v_{r_2}^{PU} \rightarrow v_{r_2}^{DO} \rightarrow v_{r_1}^{DO}\]
\[(5) \quad v_{r_2}^{PU} \rightarrow v_{r_1}^{PU} \rightarrow v_{r_1}^{DO} \rightarrow v_{r_2}^{DO}\]
\[(6) \quad v_{r_2}^{PU} \rightarrow v_{r_1}^{PU} \rightarrow v_{r_2}^{DO} \rightarrow v_{r_1}^{DO}\]

Likewise, a request $r$ and vehicle $v$ are connected in the RV graph if $r$ can be served by $v$ with maximally 5 minutes delay. With the resulting RV graph, the second step of the method is to find cliques and hence derive potential trips from them. A trip is considered feasible if every request can be picked up and dropped off by a single vehicle with a delay of less than 5 minutes. This results in a new request-trip-vehicle (RTV) graph. To lower computation times, we set a timeout of 5 seconds for the exploration of new trips.

The last step is to compute the optimal assignment of vehicles to trips. This is done through optimization of an Integer Linear Program (ILP), where an initial solution is obtained by performing a greedy assignment (maximizing the number of requests served while minimizing cost). The overall cost function to be minimized is defined as the sum of estimated delays for all requests served, plus a penalty factor $c_{ko}$ (we set $c_{ko} = 60$ minutes) for every rejected request. We solve the ILP using Gurobi Optimizer in Python with a time constraint of 10 seconds, after which the solver might return a suboptimal solution. However, this is a necessary trade-off to make in order to keep computation times stable during the simulations.

\begin{figure*}[t]
  \centering
  \includegraphics[width=0.92\linewidth]{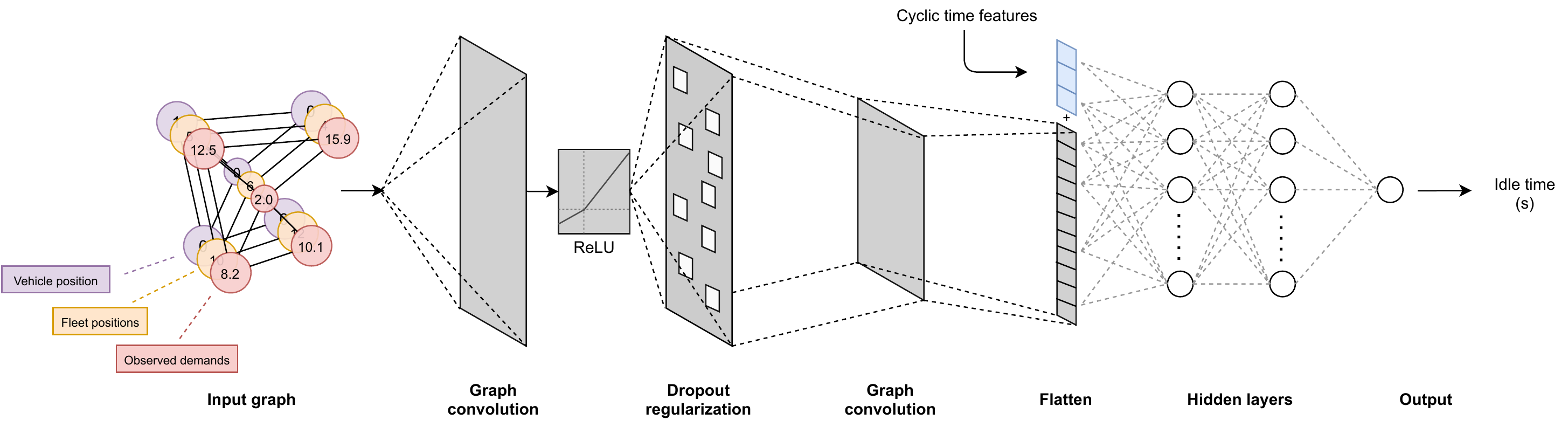}
  \caption{Graph Convolutional Network architecture for idle time prediction}
  \label{fig:gcnarchitecture}
\end{figure*}

\subsubsection{Repositioning}
\label{sec:repositioning}
After a vehicle drops off its last customer, it might be located in an area with low demands, resulting in low probabilities of serving new requests and producing larger waiting times for customers. Ideally, at any point in time, the supply (i.e. the cumulative capacity of vehicles at the vertex level) perfectly matches the demand, such that requests can be served with minimal waiting times. Since serving trip requests will always position vehicles suboptimally with regard to the demand, the repositioning process is important to ensure that the fleet will be positioned in such way that anticipated demand is covered as well as possible. In current practice, this task is often performed decentrally by individual drivers.

In this research, we decide to implement a heuristic repositioning algorithm which optimizes the repositioning decisions every minute. Our reasoning for this is that the repositioning process is not the main focus of this work. Moreover, a heuristic approach is shown to yield good results with very low computational costs \cite{clairand2019}. In the optimization task, we attempt to maximize coverage of the observed demand while minimizing travel times for repositioning movements. Here the assumption is made that the recently observed demand is a sufficiently accurate representation of the demand in the near future. First, we compose an array $X$ containing the average observed demands for every vertex in the past 60 minutes, i.e. $|X| = |N|$. This forms a sliding window of demands. $X$ is then multiplied by a parameter $H$ which is defined as the horizon (in minutes) for which the demand should be satisfied. We then perform an \texttt{argsort} operation, which produces an array of vertices which are sorted in decreasing order of observed demand. While not all idle vehicles $V_{idle} \subseteq V$ have been assigned to a vertex for repositioning, the vertex $n \in N$ with the highest remaining unserved demand is selected. The remaining idle vehicles $V_{idle}$ are then sorted based on a heuristic $h(v, n) = \frac{c_v}{t_{travel}^{v, n}}$ which divides the vehicle capacity $c_v$ (i.e. the number of seats) by the estimated travel time from the current position of $v$ to vertex $n$. The reasoning behind this is that the vehicle with the highest capacity should be assigned to the areas with the highest demand while minimizing the travel time to that demand. After computing the heuristic for all remaining idle vehicles, we assign the vehicle with the highest heuristic value to the vertex $n$ under the condition that $t_{travel}^{v, n} \leq H$. When this condition is fulfilled, we add $v$ to the list of repositioned vehicles and compute the remaining demand $x_{n}$ by subtracting the capacity of the selected vehicle. We then insert the corresponding vertex at its correct place within the array of vertices, such that it is again correctly sorted by the remaining demand. This process is repeated until all idle vehicles have been repositioned or until $\sum_{x \in X} x \leq 0$, i.e. when all demand has been satisfied.

\begin{figure*}[t]
  \centering
  \includegraphics[width=0.85\linewidth]{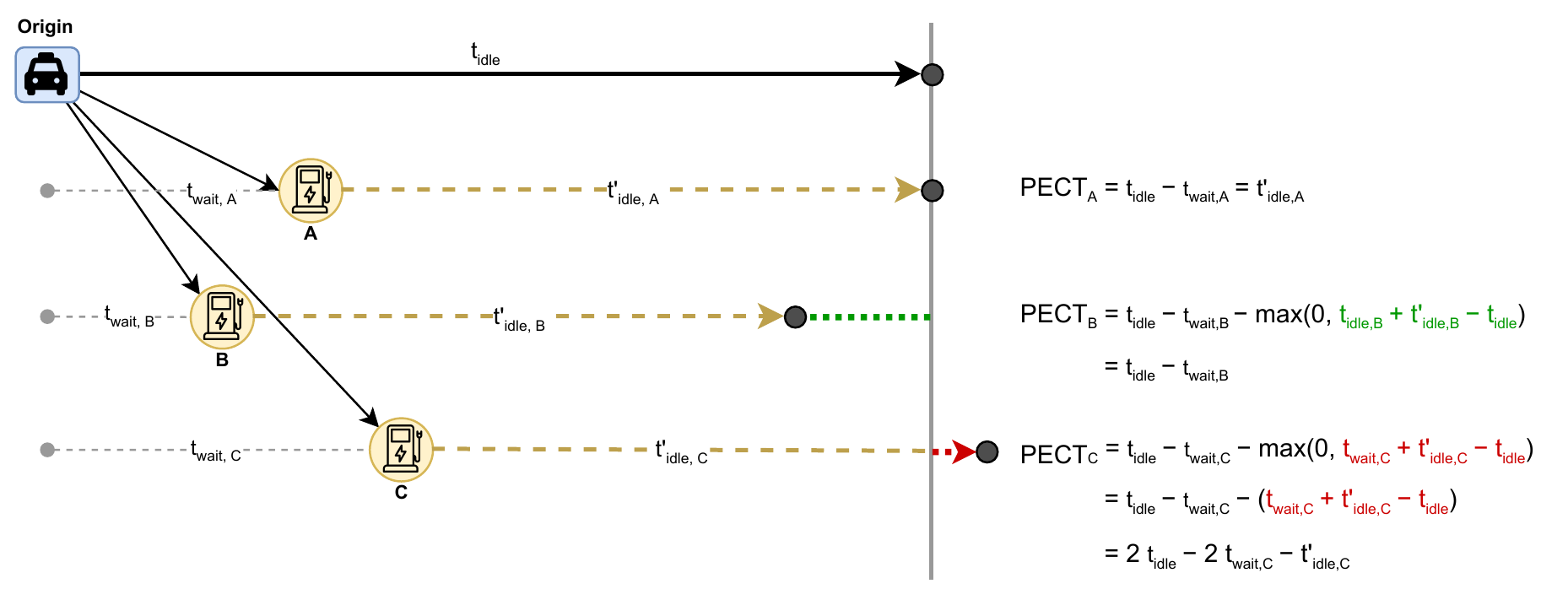}
  \caption{Schematic diagram for the comparison of rewards $R$ of charging sessions in the temporal domain}
  \label{fig:chargingcontrol}
\end{figure*}

\subsection{Charging control}
\label{sec:chargingcontrol}
The Idle Time Exploitation (ITX) method for charging control consists of multiple components which are executed sequentially during the execution of a single timestep in the simulator, as described in Section \ref{sec:simulationprocesses}. First, the idle vehicles are determined in the simulation, after which the idle times are predicted using a dedicated Graph Convolutional Network (GCN) model. This process, as well as the training procedure of the GCN, is described in Section \ref{sec:idletimeprediction}. Subsequently, a linear assignment problem is formed and solved with the Hungarian algorithm. The latter process is described in Section \ref{sec:allocation}.

\subsubsection{Idle time prediction}
\label{sec:idletimeprediction}
The essential component of the idle time-exploitation algorithm is the prediction of idle times throughout the network. The idle time for an individual vehicle can be defined as the \emph{time elapsed between the dropoff of the vehicle's last customer (i.e. moving to an idle state) and the allocation of a new customer to the vehicle}. It should be noted that the idle time prediction model is only trained on samples where the vehicle becomes idle after having served all of its customers. Hence, repositioning can occur during the idle time period but the end of a repositioning action is not regarded as the starting time for the measurement of an idle time sample. The reasoning for this is that we aim to enable the model to learn directly from the system dynamics, including all non-charging related processes that run inside the simulator. To illustrate, when a vehicle becomes idle after having served all of its passengers, we measure the current timestep in the simulator and assign it to the vehicle object. When a new customer is not immediately assigned to the vehicle, it will be considered for repositioning. Once the vehicle gets a new customer assigned to it during or after repositioning, the simulator time is again measured and the initial time measurement is subtracted from this new measurement. This way, a new idle time sample is obtained and added to the training batch, such that it can be used to train the idle time prediction model.

Given the complexity of the dispatching and repositioning processes, it is arguably crucial that the model is provided with enough information, such that it can infer the spatiotemporal relations between these processes and the idle times that are measured. We argue that the idle time is both dependent on the individual vehicle state (capacity, current location) and the state of the environment (spatial distribution of other vehicles' capacities, current time, observed demands at all vertices). We decide to use a one-hot encoding to represent the current location of the vehicle, i.e. a zero-filled array with length $|N|$ where the current vertex is set to 1. Additionally, we represent the fleet positions (i.e. spatial distribution of other vehicles) by an identically sized array where the total available seats is aggregated per vertex in the road network. Lastly, another layer of input data is created using a sliding window of observed demands. Here, the aggregated (mean) demand for every vertex is computed for the past 60 minutes in the simulator. This yields another array of length $|N|$. These three layers of input arrays are then concatenated into a $3\times|N|$ matrix which serves as the main input for the idle time prediction model. The computed idle time sample serves as the output. Together with the $3\times|N|$ input matrix and an array of temporal features, they form a data sample which is used to train, validate and test the model.

To enhance the model's ability to interpret the relations between vertices, we employ the Graph Convolutional Network (GCN) technique proposed by Kipf and Welling \cite{kipf2017}. In the model structure, we implement two GraphConv layers using the Spektral library \cite{spektral}, and in between these layers we use a Dropout layer (with a dropout rate of 0.5) to reduce overfitting. We place the ReLU activation function after the GraphConv layers, after which we flatten the tensor and concatenate it with the array of time features (hour, minute and weekday). Subsequently, we implement two dense layers (with ReLU as activation function) which feed into a single neuron that outputs the idle time in seconds. During training, we use the Mean Squared Error (MSE) as the loss function. In order to increase predictive performance, we perform hyperparameter tuning on the GCN. The most important hyperparameters that should be tuned are: the \textbf{number of filters} (in the GraphConv layers), the \textbf{number of neurons} (divided among the two fixed dense layers) and the \textbf{learning rate}.

During the idle time training runs, we assume a hypothetical scenario where all vehicles are fully charged and where no energy is consumed during driving. As a consequence, vehicles do not need to charge, and therefore the charging decision process is disregarded upon training of the idle time prediction model. This is to prevent feedback loops from destabilizing the system and inducing highly non-stationary behaviour into the model, thus making it harder for the model to reliably predict across different scenarios and environments. Such behaviour is an unavoidable consequence of using model outputs to (either directly or indirectly) determine a charging policy, as it will influence the idle times and actions taken in future iterations. Over time, this positive feedback loop will move the system away from an equilibrium state. Therefore, we let the model learn from the system dynamics and policies that are induced by the two processes that are unconditionally present: dispatching and repositioning.

\subsubsection{Charger-vehicle allocation}
\label{sec:allocation}
To determine how the idle vehicles in the fleet can most optimally exploit the predicted idle time, an assignment needs to be performed. Suppose that $V_{idle} \subseteq V$ is the subset of vehicles that is currently idle, i.e. not serving any passengers, not dispatching and not repositioning or charging. We define $S$ to be the set of chargers that are present inside the simulated environment. We then form a bipartite graph $G = (V_{idle}, S, T)$, with $T$ being the set of edges that connect an idle vehicle $v \in V_{idle}$ to a charging station $s \in S$ if the algorithm finds that the predicted idle time $t_{idle, v}$ can be exploited for charging $v$ at $s$. $t_{idle, v}$ is obtained by performing an idle time prediction for the current location of $v$ and at the current time, while $t^{*, idle}_{v}$ represents another idle time prediction which takes as input the updated location and time after the (hypothetical) trip to $s$ has been completed. For every combination of $v \in V_{idle}$ and $s \in S$, we compute the Potential Effective Charging Time (PECT) and assign it to the edge $T_{v, s}$. This yields a complete bipartite graph. The PECT for edge $e(v, s)$ can be computed using Equation \ref{eq:pect}. The PECT equation was designed such that it is negative when the waiting time is larger than the predicted idle time, i.e. when no idle time can be exploited. The waiting time $t^{wait}_{v, s}$ is defined as the maximum of the travel time $t^{travel}_{v, s}$ and the expected queuing time $t^{queue}_{s}$, i.e. $t^{wait}_{v, s} = max\{t^{travel}_{v, s}, t^{queue}_{s}\}$. Also, the equation includes a penalty in case that the sum of the waiting time $t^{wait}_{v, s}$ and the predicted idle time $t^{*, idle}_{v}$ at the new location are larger than the idle time at the original location $t^{idle}_{v}$. We argue that this is a suitable way to penalize a potential assignment $e(v, s)$ if it requires $v$ to move to a less attractive area $s$ (i.e. lower likelihood of picking up customers) to charge. This could lead to situations where vehicles have more difficulties finding a customer as a consequence of their charging decisions, which could negatively influence operational performance. A compromise is therefore embedded into the PECT computations such that, after the charging session, the vehicle has time to reposition to an area with higher demand. An example schematic of idle time predictions and corresponding PECT calculations is visible in Figure \ref{fig:chargingcontrol}.

\begin{equation}
    PECT(v, s) = t^{idle}_v - t^{wait}_{v, s} - max\{0, t^{wait}_{v, s} + t^{*, idle}_{v} - t^{idle}_v\}
    \label{eq:pect}
\end{equation}

After computing the PECT values for all combinations of $v \in V_{idle}$ and $s \in S$ and assigning them to the edges $e(v, s) \in T$. Then the edges are removed where $PECT < 0$, such that only the potential assignments remain where idle time can be exploited for charging. An exact solution to the resulting assignment problem (i.e. finding an assignment where the sum of PECT values is maximal) can then be found in $O(n^3)$ polynomial time using the Hungarian algorithm \cite{jonker1988}. Subsequently, we store the assignments in the form of tuples which contains the vehicle ID, the charger ID and the PECT. If the set of assigned vehicles $V_{assigned} \subseteq V_{idle}$ is not empty, we derive a new set $V'_{idle}$ by subtracting the assigned vehicles from the initial set of idle vehicles, i.e. $V_{idle} - V_{assigned}$. We then repeat the previous steps: first we recompute the PECT values for the combinations of $v \in V'_{idle}$ and $s \in S$ (since the queuing times have changed since the first iteration), after which we remove the edges where $PECT < 0$ and solve the remaining assignment problem with the Hungarian algorithm. We perform multiple iterations of this process, until either (1) there are no edges $e \in T$ for which $PECT > 0$ or (2) the remaining set of idle vehicles is empty $V'_{idle} = \emptyset$. Through iteration, we make sure that the optimization problem is tractable, and that we can assign multiple vehicles to a charging station. This yields a solution that could be suboptimal but is obtained in polynomial time. The algorithm is described in pseudocode in Algorithm \ref{alg:assignment}.

\begin{algorithm}[h]
\SetAlgoLined
\KwIn{Idle time prediction model, set of vehicles $V$, set of charging stations $S$}
\KwOut{Vehicle-charger mapping $(v, s, PECT(v, s))$ for all optimized assignments}
 \textbf{Initialize} set of idle vehicles $V_{idle} \subseteq V$; $V'_{idle} = V_{idle}$\;
 \textbf{Initialize} empty set of assignments $A = \{\emptyset\}$\;
 \textbf{Initialize} a bipartite graph $G = (V, S, T)$\;
 \textbf{Predict} the idle times $t^{idle}_v$ for $\forall v \in V$\;
 \While{$\exists (v,s) \in T$ and $PECT(v, s) > 0$ and $|V'_{idle}| > 0$}{
     \For{$v \in V'_{idle}$}{
         \For{$s \in S$}{
            \textbf{Compute} $PECT(v, s) = t^{idle}_v - t^{wait}_{v, s} - max\{0, t^{wait}_{v, s} + t^{*, idle}_{v} - t^{idle}_v\}$\;
            \eIf{$PECT(v, s) > t_{min}$}{
                \textbf{Assign} $PECT(v, s)$ to edge $e(v, s)$ in $G$\;
            }{
                \textbf{Remove} edge $e(v, s)$ from $G$\;
            }
        }
     }
     \textbf{Solve} assignment problem using Hungarian algorithm\;
     \textbf{Append} optimal assignments to $A$\;
     \textbf{Set} $V'_{idle} = V_{idle} - V_{assigned}$\;
 }
 \caption{Solving charger-vehicle assignment}
 \label{alg:assignment}
\end{algorithm}

\section{Experimental Setup}
\label{sec:experimentalsetup}
\subsection{Hyperparameter tuning}
\label{sec:hyperparametertuning}
As mentioned in Section \ref{sec:idletimeprediction}, the idle time model is trained and validated on a dataset of idle times which is generated during an independent simulation run. This simulation run is performed on a full month of trip request data from November 2015. To prevent the occurrence of feedback loops, we make sure that all vehicles are fully charged and that no energy is consumed during driving. Therefore, no charging takes place during these simulation runs. After a new customer is assigned to a vehicle, a new row is written to the dataset containing: the \emph{vehicle position upon becoming idle}, the \emph{spatial distribution of the fleet upon becoming idle}, the \emph{observed demands in the last hour before becoming idle} and the cyclic time features \emph{hour}, \emph{minute} and \emph{weekday}. Finally, we add the dependent variable, i.e. the observed \emph{idle time} (in seconds).

\begin{figure}[h]
\centering
\begin{blockarray}{cccccccc}
    & \BAmulticolumn{6}{c}{\text{\# of neurons}} & \\
    & 128 & 256 & 512 & 1024 & 2048 & 4096 & \\[3px]
    \begin{block}{c[cccccc]c}
        32 & $L_{32}^{128}$   & $L_{32}^{256}$ & $L_{32}^{512}$ & $L_{32}^{1024}$ & \ $L_{32}^{2048}$ & $L_{32}^{4096}$ & \multirow{4}{*}{\rotatebox{-90}{\text{\# of filters}}} \\[5px]
        64 & $L_{64}^{128}$   & $L_{64}^{256}$ & $L_{64}^{512}$ & $L_{64}^{1024}$ & \ $L_{64}^{2048}$ & $L_{64}^{4096}$ & \\[5px]
        128 & $L_{128}^{128}$   & $L_{128}^{256}$ & $L_{128}^{512}$ & $L_{128}^{1024}$ & \ $L_{128}^{2048}$ & $L_{128}^{4096}$ & \\
    \end{block}
\end{blockarray}
\caption{Matrix of grid search for GCN architecture}
\label{fig:gridsearchgcn}
\end{figure}

After performing the simulation run, the dataset of idle time observations is obtained. This dataset is then randomly shuffled and split into a 70\% training set, 10\% validation set and a 20\% test set. We then perform a multitude of training rounds (for 200 epochs, using the Adam optimizer and with reference value $\alpha = 0.0001$) with varying model architectures (i.e. the number of neurons and number of filters in the GraphConv layers) and evaluate the performance on the validation set using the Mean Absolute Error (MAE) metric. The resulting grid search will produce a table of MAE values $L^n_m$ for $n$ neurons (divided equally among the two dense layers) and $m$ filters, as visible in Figure \ref{fig:gridsearchgcn}. The combination $n, m$ with the lowest MAE score is used for the next round of hyperparameter tuning, as well as training of the final model.

\begin{figure}[h]
\centering
\begin{blockarray}{cccccccc}
    & \BAmulticolumn{6}{c}{\text{Learning rate $\alpha$}} & \\
    & $10^{-1}$ & $10^{-2}$ & $10^{-3}$ & $10^{-4}$ & $10^{-5}$ & $10^{-6}$ \\
    \begin{block}{c[cccccc]c}
         & $L_{10^{-1}}$   & $L_{10^{-2}}$ & $L_{10^{-3}}$ & $L_{10^{-4}}$ & $L_{10^{-5}}$ & $L_{10^{-6}}$ \\
    \end{block}
\end{blockarray}
\caption{Array of grid search for learning rate $\alpha$}
\label{fig:gridsearchlr}
\end{figure}

With the optimized architecture of the model, we try to optimize the learning rate $\alpha$ which determines the step size at each iteration of the training process. The aim is to find the right balance between fast convergence and stability of the loss function. This time, we evaluate both the $R^2$ score and Mean Absolute Error (MAE) during a grid search of $\alpha$ values from $10^{-6}$ to $10^{-1}$. The configurations in this grid search are displayed in Figure \ref{fig:gridsearchlr}. Again, for all configurations the training is performed for 200 epochs with the Adam optimizer. We select the value of $\alpha$ where the $R^2$ score is highest and the MAE is lowest.

Finally, we train the model with the optimized hyperparameter settings on the complete training set for 1,000 epochs. An evaluation of the trained model on the test set will provide a conclusive performance measure for the idle time prediction model.

\subsection{Baseline charging strategies}
\label{sec:baselines}
\begin{table*}[h]
\resizebox{\linewidth}{!}{%
\begin{tabular}{l|lll}
\textbf{Baseline strategy}                    & \textbf{When to charge?}     & \textbf{Where to charge?}                    & \textbf{For how long?} \\ \hline
Quick Charge, Nearest Station (QN)   & SoC \textless 10\% & Nearest charging station           & Until SoC = 70\%   \\
Quick Charge, Available Station (QA) & SoC \textless 10\% & Nearest available charging station & Until SoC = 70\%   \\
Full Charge, Nearest Station (FN)    & SoC \textless 10\% & Nearest charging station           & Until SoC = 99\%   \\
Full Charge, Available Station (FA)  & SoC \textless 10\% & Nearest available charging station & Until SoC = 99\%   \\
Overnight Quick Charge (OQ)              & Time between 01:30 and 06:30 & Nearest available charging station & Until 06:30 or SoC = 70\% \\
Overnight Full Charge (OF)              & Time between 01:30 and 06:30 & Nearest available charging station & Until 06:30 or SoC = 99\%
\end{tabular}}
\caption{Description of baseline strategies}
\label{tab:baselines}
\end{table*}

To evaluate the proposed method, we perform the simulator runs under identical conditions with a selection of baseline charging strategies. For these baseline strategies, various levels of complexity and information provision are considered. For instance, one of the simplest baseline methods is to charge every vehicle overnight, or to perform a full charge at the charging station which is closest to the current position of the vehicle. When one knows the availability and queuing times of the charging infrastructure, a more intelligent baseline can be established. All baselines are described in Table \ref{tab:baselines}.

The first two baselines that we use involve a quick charge, i.e. when a vehicle has less than 10\% of its charge left, the battery will be charged until the tapering point is reached at 70\%. This way, we avoid the problem of the charging rate decreasing and the State-of-Charge (SoC) slowly reaching the asymptote of 100\%, as this is relatively time-inefficient. The \textbf{Quick Charge, Nearest Station} (QN) baseline follows this principle and selects the charging station that is nearest to the vehicle, independent of the availability and estimated queuing times of the charging stations. Therefore, the possibility arises that vehicles are distributed in an unbalanced manner among the charging infrastructure. Together with the \textbf{Full Charge, Nearest Station} (FN) baseline, we argue that this baseline can be considered as the least intelligent and most comparable to a real-world scenario where drivers make egoistic charging decisions and are not aware of the real-time status of the infrastructure.

In contrast to the QN and FN baselines, the \textbf{Quick Charge, Available Station} (QA) and \textbf{Full Charge, Available Station} (FA) baselines assume that more information on infrastructure and fleet status is available to the driver. Similarly to QN and FN, a vehicle is charged when its SoC is below 10\%. However, in QA and FA the charging location is decided by the availability of the charging stations. The travel and queuing times are computed for every charging station based on their current status. Additionally, the estimated energy consumption will be computed in order to ignore charging stations that cannot be reached. Consequently, every vehicle obtains a ranking of charging stations. The most attractive charging station (i.e. with lowest travel plus queuing times) is then selected as the location to charge. The QA baseline performs a quick charge until the SoC reaches 70\%, while the FA baseline performs a (near) complete charge up to a SoC of 99\%.

We also implement a baseline for \textbf{Overnight Quick Charge} (OQ) and \textbf{Overnight Full Charge} (OF). After performing an analysis of the trip request dataset, we conclude that the number of trip requests decreases drastically between 01:30 and 06:30 in the early morning \cite{tlctaxidata}. This leaves a five hour window to gradually charge the fleet. Based on the number of chargers $|S|$ in the simulation, the $|S|$ vehicles with the lowest SoC are sent to these chargers. The vehicle with the lowest SoC is assigned to the charging station with the lowest expected travel plus queuing time. Subsequently, once a charger becomes vacant again, the vehicle with the then-lowest SoC will be assigned to the corresponding charging station. This process will continue until either all vehicles have been charged, or until the time is 06:30 AM.

\subsection{Experimental design}
\subsubsection{Settings and parameters}
\label{sec:settingsandparameters}
By performing extensive simulator runs, we aim to evaluate the effectiveness of the proposed method in comparison to the baseline methods. For the experiments, a period of one week was selected, from Monday November 2 until Sunday November 8, 2015. This period was selected based on the criteria that no public holidays, events or special weather conditions take place. Therefore, the simulations are arguably performed based on normal taxi movement patterns in the city of New York. All vehicles have an initial charge which is randomly sampled from a continuous uniform distribution between 50\% and 100\% SoC. Similarly, the initial location of the vehicles is determined by a random sampling with equal probabilities of the vertices $N$ in the road network. An initialization period of three simulated days is used to prepare the simulation for the actual experiments: here no energy is consumed and charging does not occur yet. This way, we let the fleet adapt to the demand ahead of the experiments, such that the starting conditions do not influence the results. Hence, when the time reaches November 2, 0:00, the experiments will commence and the metrics will be saved.

We perform the experiments with a fleet of 3,000 Nissan Leaf hatchbacks, 2,000 Tesla Model 3 LR sedans and 1,000 Nissan e-NV200 vans. Overall, we therefore simulate a ridepooling system with a fleet size $|V| = 6000$. In the main experimental runs, using the process described in Section \ref{sec:chargermodeling}, $|S| = 500$ chargers with a supply power of 72 kW are randomly placed on the road network. In additional experiments, the impact of the charging infrastructure on the profitability of the ridepooling service is evaluated. Therefore, we run the simulations under identical circumstances with $|S|$ ranging from 100 to 900 in step sizes of 200. The horizon $H$ for the repositioning algorithm (see Section \ref{sec:repositioning}) is set to 30 minutes. The minimum charging time $t_{min}$ for the idle time exploitation algorithm is set to 5 minutes.

The experiments were performed on computing resources provided by the Swedish National Infrastructure for Computing (SNIC) at Chalmers Centre for Computational Science and Engineering (C3SE) \cite{c3se}. The computing nodes that were utilized consist of a Nvidia Tesla T4 GPU and Intel Xeon Gold 6226R CPU.

\subsubsection{Assessment framework}
The metrics that are used to evaluate the effectiveness of the proposed methodology can be divided in two categories. First of all, we assess \emph{resource efficiency}, which involves the cost implications for the operator as well as the burden on the electricity grid, as this would signify eventual improvements regarding the environmental impact of the ridepooling operations. Additionally, we consider \emph{transport efficiency}, which concerns the delays and comfort experienced by customers of the ridepooling service. The full list of metrics is defined as follows:

\begin{enumerate}
    \item Resource efficiency
    \begin{itemize}
        \item The \textbf{reward} (in \$) which is achieved by the operator. This forms the central objective function of this research, since we argue that the other metrics greatly influence this reward function and form a function where the weighted components are represented by their estimated real-world importance to the overall revenue model of the ridepooling service. The reward function is explained in more detail in Figure \ref{sec:rewardfunction}.
        \item The average \textbf{SoC of the fleet} (in \%) over time, which is defined as the average State-of-Charge (or battery percentage) across the entire fleet of vehicles.
        \item The \textbf{charger occupancy rate} (in \%), which is the percentage of chargers in the network that is occupied. This metric enables us to observe whether the charging loads are spread out over time.
        \item The \textbf{power drawn from the grid} (in MW) over time and space. This is strongly related to the previous metric, but gives some more insight into the actual loads that are drawn from the grid and the spatial distribution of loads.
        \item The \textbf{energy consumed per on-time request} (in kWh). This metric is computed by dividing the total energy consumption of the fleet by the number of requests that were served on-time (i.e. within 5 minutes delay). This helps to evaluate whether the charging control functions efficiently in assigning vehicles to charging stations with low travel and queuing times while mitigating increased idle times after the charging sessions. This would arguably lead to lower overall energy consumption in the long run.
    \end{itemize}
    \item Transport efficiency
    \begin{itemize}
        \item The average \textbf{delay} (in minutes) experienced by customers. To compute the delay for a single trip, we first compute the interval between the time at which the customer is delivered to the final destination and the time at which the request was received. Then we subtract this by the direct waiting and direct travel time (i.e. the time it would take to serve the request without having to share the vehicle with other customers).
        \item The \textbf{on-time rate} (in \%), which is defined as the percentage of requests which is served with less than 5 minutes of delay.
        \item The average \textbf{number of customers per vehicle}, which can be regarded as an indicator of the comfort level experienced by customers. It is preferred when customers spread out more efficiently across the fleet.
    \end{itemize}
\end{enumerate}

\subsubsection{Runtime measurements}
To evaluate whether our proposed method is able to run in real-time, i.e. with a runtime that is consistently lower than 60 seconds (the time resolution of the simulations and decision control), we perform simulations and measure the runtimes of the charging control algorithm. We perform the simulations on a single day, Tuesday April 21 of 2015 and vary the fleet size from 5,000 to 7,000 in steps of 1,000 vehicles. Please note that we maintain the same ratio of vehicle types in the fleet and use identical parameters as in Section \ref{sec:settingsandparameters}.

\section{Results \& Discussion}
\label{sec:results}
\subsection{Model training and validation}
\begin{figure}[h]
  \centering
  \includegraphics[width=\linewidth]{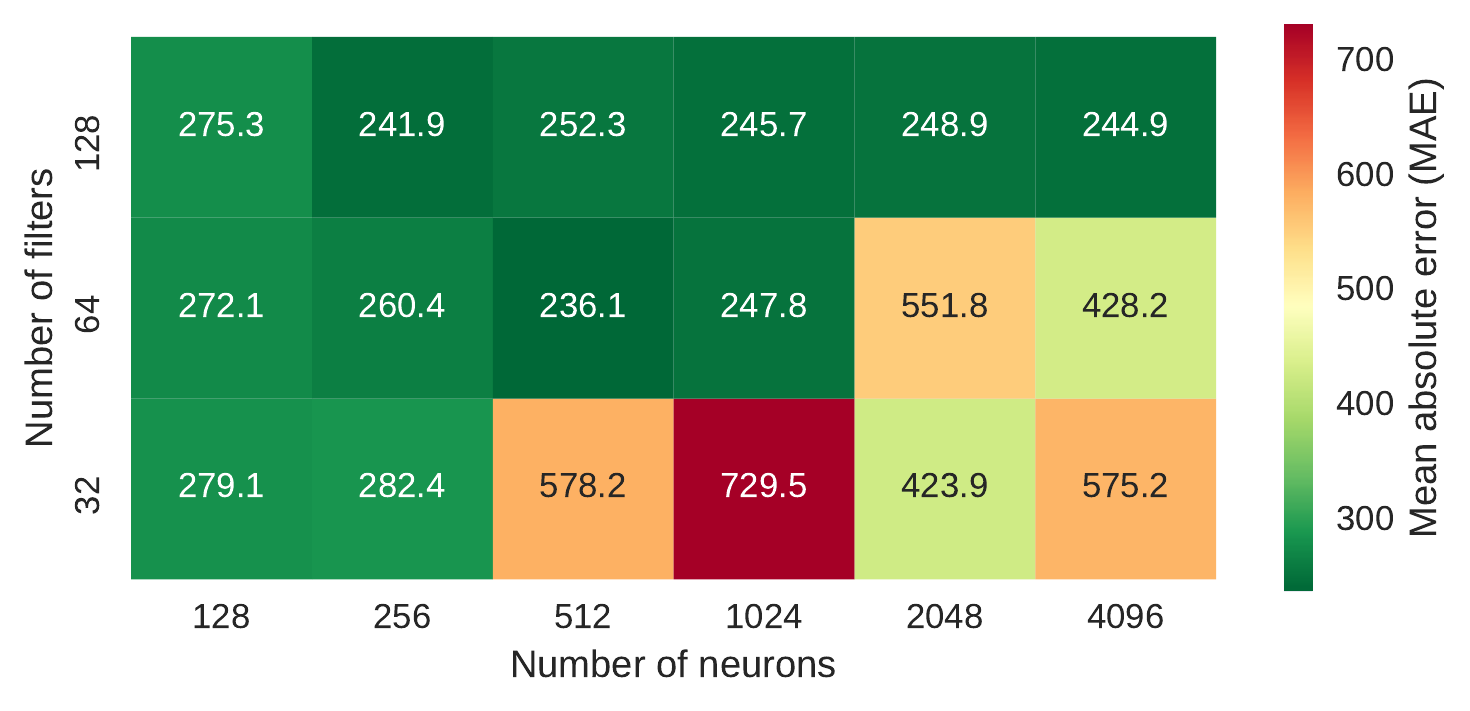}
\caption{Results of grid search with differing number of GCN filters and number of neurons in dense layers}
\label{fig:resultstuning1}
\end{figure}

First, following the methodology of Section \ref{sec:hyperparametertuning}, we perform the training and validation rounds needed to optimize the hyperparameters of the idle time prediction model. For all configurations, a new model was trained for 200 epochs on the training set using the Adam optimizer. The results of the first grid search, aimed at finding the optimal number of filters in the GCN and neurons in the dense layers, are visible in Figure \ref{fig:resultstuning1}. From the results, it emerges that finding the right balance the two hyperparameters yields an optimal result, but generally it seems like a relatively low number of neurons and high number of filters will provide better results. This could be explained by the fact that the model could slightly underfit when a smaller network architecture network is used, whereas an overly complex architecture might lead to overfitting. The configuration of $64$ filters and $512$ neurons (spread evenly among two dense layers, i.e. a $256-256$ configuration) performs best on the validation set, with a Mean Absolute Error of $236.1$ seconds. Therefore, this configuration is selected for the next grid search.

\begin{figure}[h]
  \centering
  \includegraphics[width=\linewidth]{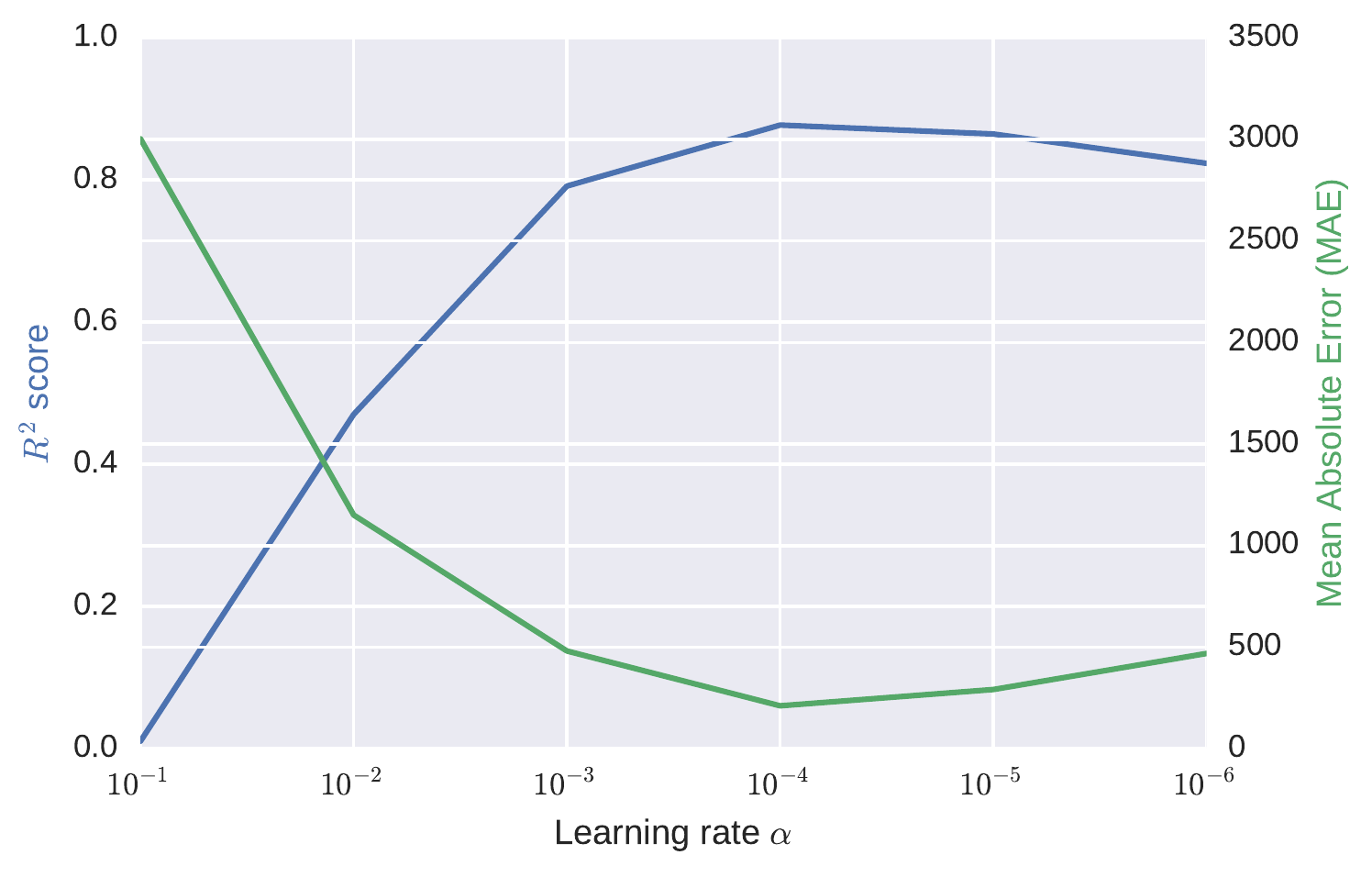}
  \small
\begin{tabular}{l|llllll}
\textbf{Learning rate} & $10^{-1}$   & $10^{-2}$   & $10^{-3}$   & $10^{-4}$   & $10^{-5}$   & $10^{-6}$   \\ \hline
MAE                    & 3003.1 & 1150.7 & 481.23 & 211.37 & 236.11 & 469.10 \\
$R^2$ score               & 0.0105 & 0.4703 & 0.7914 & 0.8772 & 0.8647 & 0.8234
\end{tabular}
\caption{Results of grid search with differing learning rates $\alpha$ on validation set}
\label{fig:resultstuning2}
\end{figure}

The results of the second grid search, aimed at finding the optimal learning rate $\alpha$, are visible in Figure \ref{fig:resultstuning2}. It becomes apparent that the performance on the validation set (i.e. both the $R^2$ and MAE) drastically improves as the learning rate decreases towards the optimum of $10^{-4}$. The fact that relatively high values of $\alpha$ yield bad results is likely caused by the risk of learning suboptimal weights too quickly, which might destabilize the training process. At the optimum of $\alpha = 10^{-4}$, the MAE equals $211.37$ while the $R^2$ score equals $0.8772$.  With an even smaller learning rate, the performance worsens again. This could be explained by the fact that small learning rates may slow down the training too much or cause the training process to get stuck, enabling the model to learn less over a period of 200 epochs. Hence, we select the learning rate $\alpha = 10^{-4}$ for the final training round, as it provides a good balance between fast and stable learning.

Finally, the idle time prediction model (with 64 filters, a $256-256$ dense layer architecture and a learning rate $\alpha = 10^{-4}$) is trained for 1,000 epochs on the training set with its definitive configuration. We then evaluate the performance on the (so far unused) test set, which produces an MAE of $210.86$ seconds and an $R^2$ score of $0.8792$. The idle time prediction model is then saved, such that it can be loaded into memory upon initialization of the simulator.

\subsection{Simulation runs}
\begin{table*}[h]
\centering
\resizebox{\linewidth}{!}{%
\begin{tabular}{llllll}
\toprule
Charging strategy & Reward (\$) & Delay (min.) & On-time rate (\%) & Customers/veh. & Energy/on-time req. (kWh) \\ \hline
\textbf{\textit{Idle Time Exploitation (ITX)}} & \textbf{1,487,347} & \textbf{0.805} & \textbf{95.78} & 0.611 & \textbf{0.975} \\
\textit{(B) Quick Charge, Nearest Station (QN)} & 235,088 (-84.19\%) & 19.30 (+2,298\%) & 61.85 (-33.93\%) & 1.195 (+95.58\%) & 1.183 (+21.33\%)\\
\textit{(B) Quick Charge, Available Station (QA)} & 1,417,756 (-4.679\%) & 0.886 (+10.06\%) & 89.30 (-6.478\%) & \textbf{0.607} (-0.654\%) & 1.025 (+5.128\%)\\
\textit{(B) Full Charge, Nearest Station (FN)} & 45,554 (-96.94\%) & 27.19 (+3,278\%) & 56.93 (-38.84\%) & 1.322 (+116.3\%) & 1.239 (+27.08\%)\\
\textit{(B) Full Charge, Available Station (FA)} & 1,412,909 (-5.005\%) & 2.348 (+191.7\%) & 84.44 (-8.443\%) & 0.712 (+16.53\%) & 1.032 (+5.846\%)\\
\textit{(B) Overnight Quick Charge (OQ)} & 1,086,435 (-26.95\%) & 3.598 (+347.0\%) & 80.85 (-14.92\%) & 0.769 (+25.86\%) & 1.054 (+8.102\%)\\
\textit{(B) Overnight Full Charge (OF)} & 945,993 (-36.40\%) & 6.236 (+674.7\%) & 75.59 (-20.19\%) & 0.905 (+48.12\%) & 1.045 (+7.179\%)\\
\bottomrule
\end{tabular}}
\caption{Performance comparison of charging strategies (best-performing denoted in bold) for $|S| = 500$ and $|V| = 6000$}
\label{tab:results}
\end{table*}

\begin{figure*}[h]
  \begin{subfigure}[b]{0.33\textwidth}
    \includegraphics[width=\textwidth]{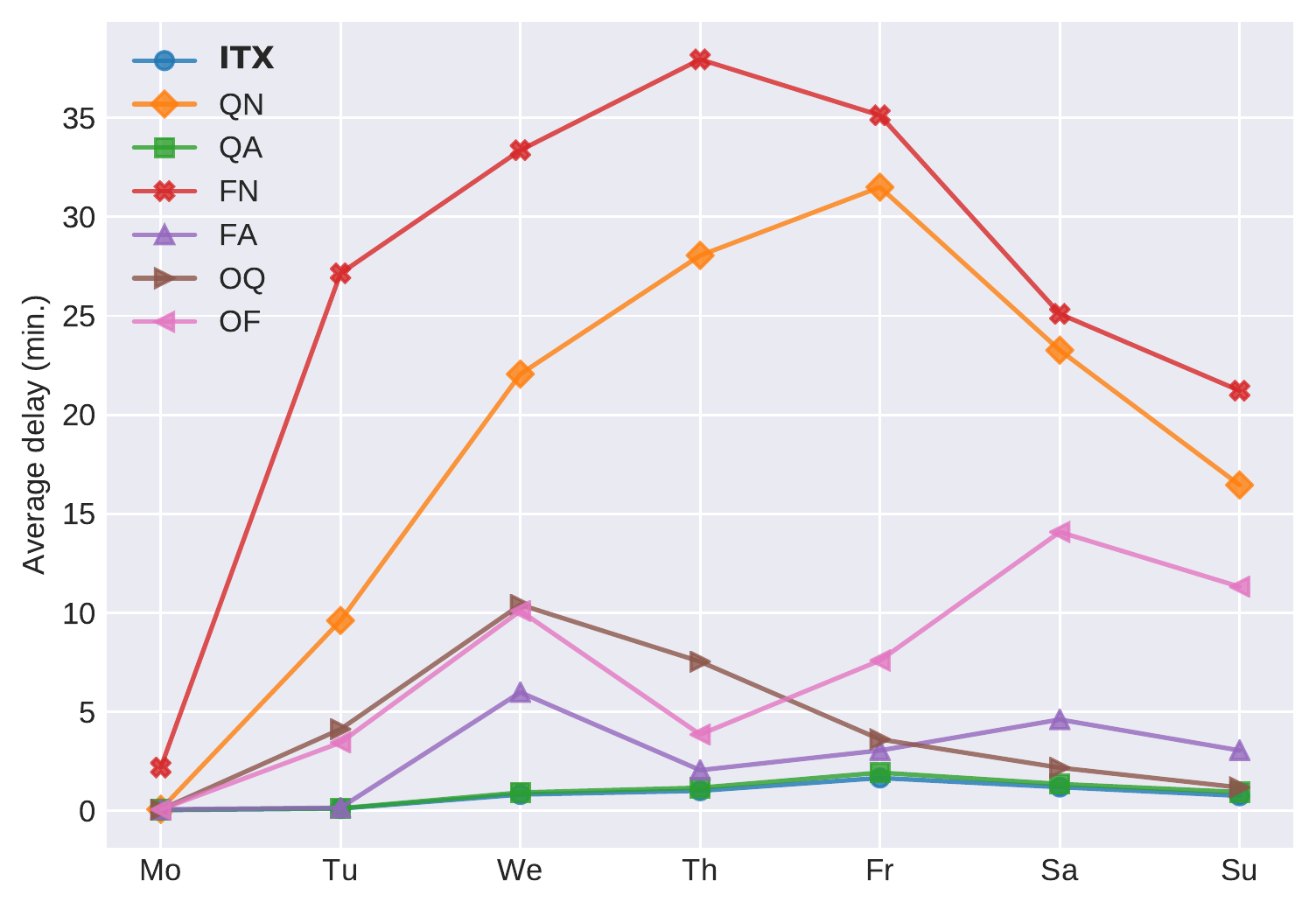}
    \caption{Delay}
    \label{fig:delay}
  \end{subfigure}
  \begin{subfigure}[b]{0.33\textwidth}
    \includegraphics[width=\textwidth]{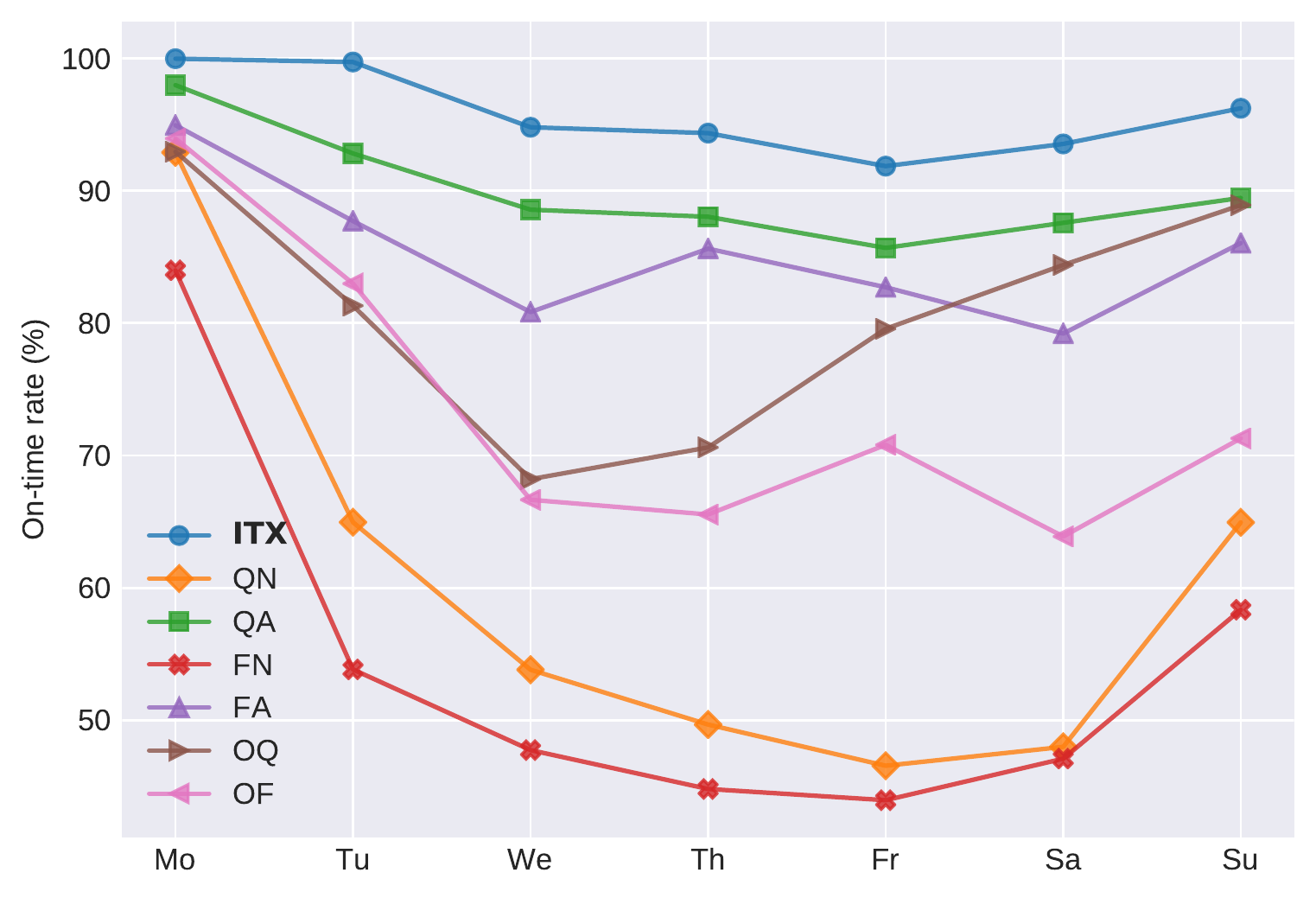}
    \caption{On-time rate}
    \label{fig:ontimerate}
  \end{subfigure}
  \begin{subfigure}[b]{0.33\textwidth}
    \includegraphics[width=\textwidth]{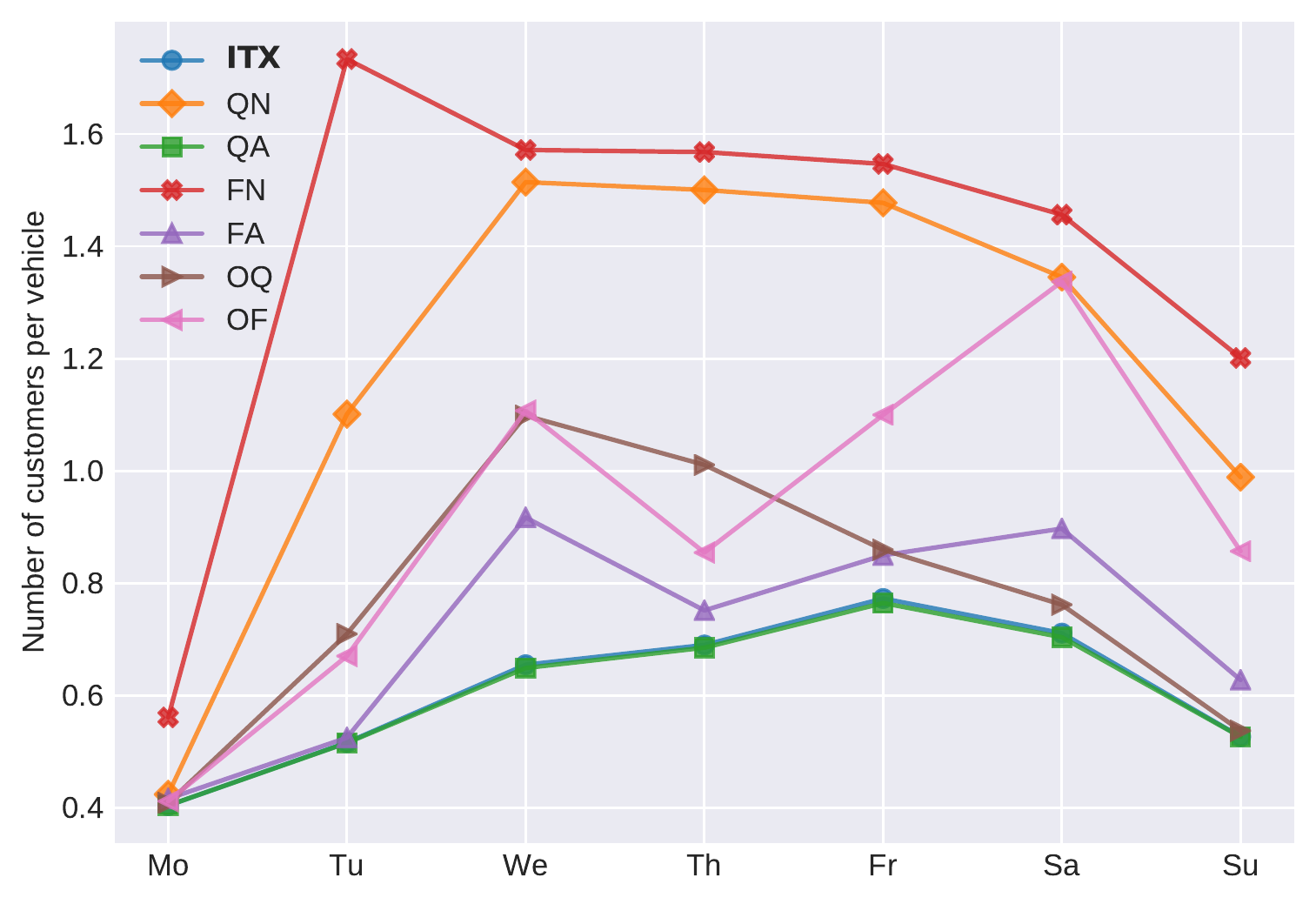}
    \caption{Number of customers per vehicle}
    \label{fig:customerspervehicle}
  \end{subfigure}
  \begin{subfigure}[b]{0.33\textwidth}
    \includegraphics[width=\textwidth]{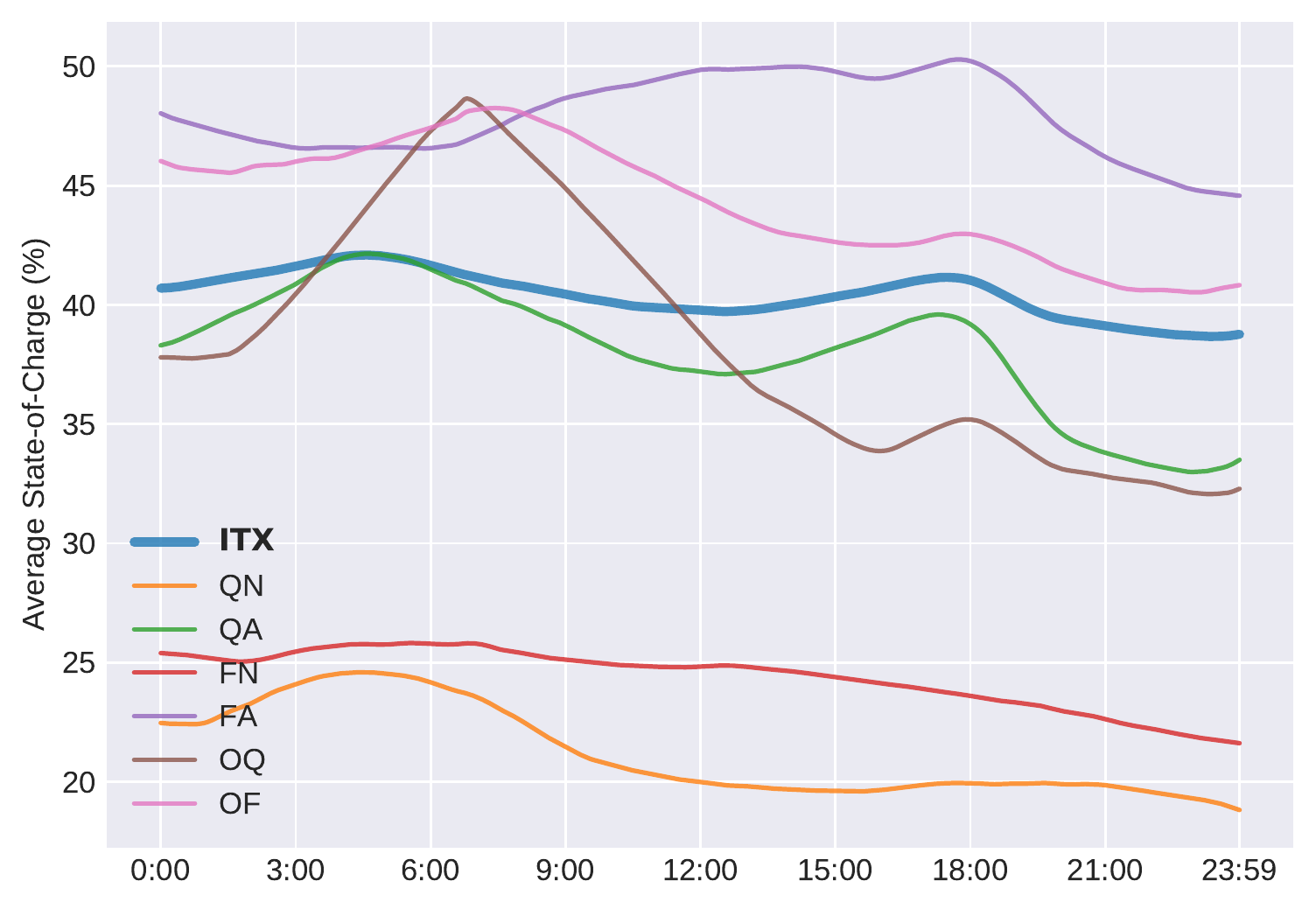}
    \caption{State-of-Charge}
    \label{fig:stateofcharge}
  \end{subfigure}
  \begin{subfigure}[b]{0.33\textwidth}
    \includegraphics[width=\textwidth]{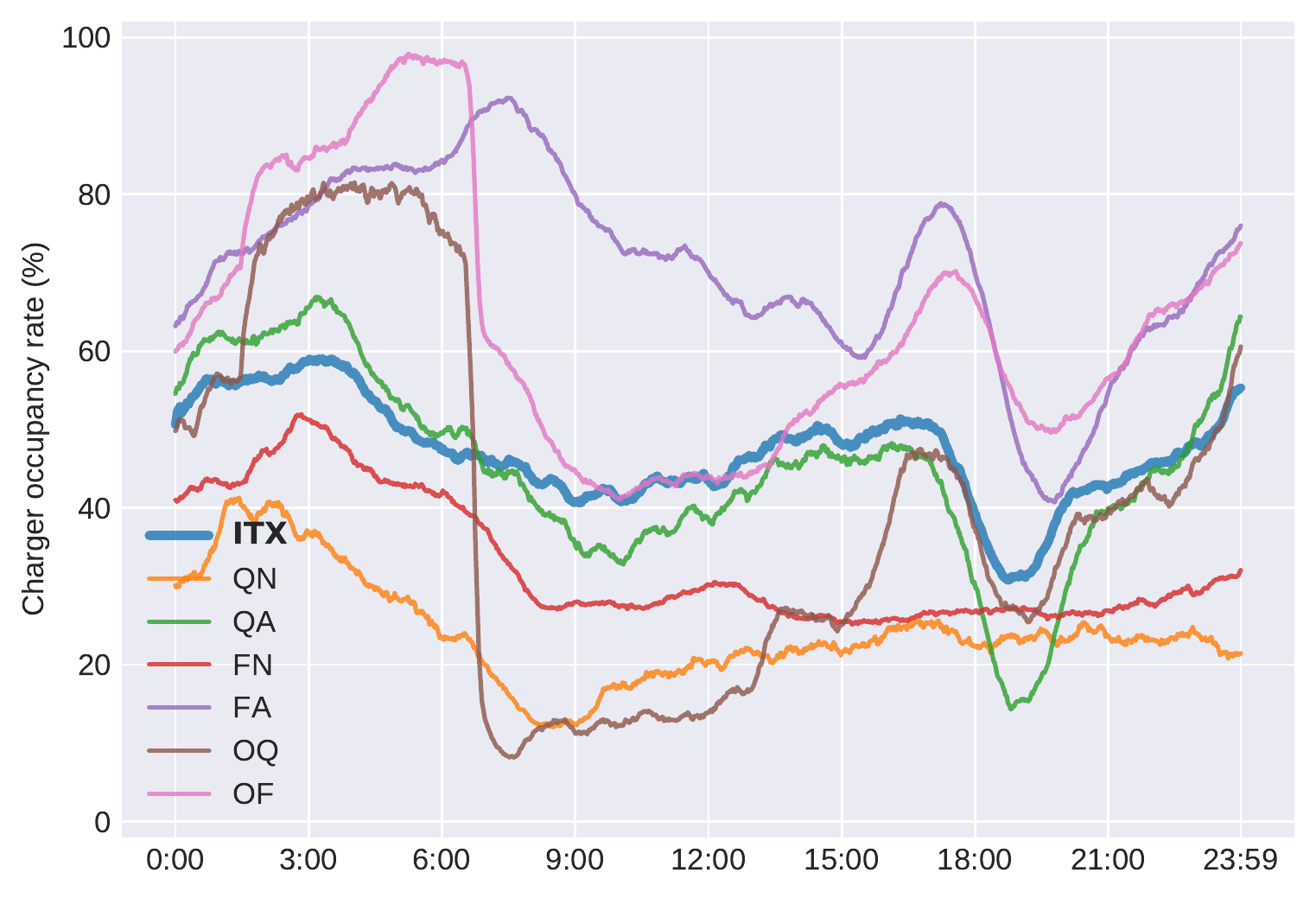}
    \caption{Charger occupancy rate}
    \label{fig:chargeroccupancy}
  \end{subfigure}
  \begin{subfigure}[b]{0.33\textwidth}
    \includegraphics[width=\textwidth]{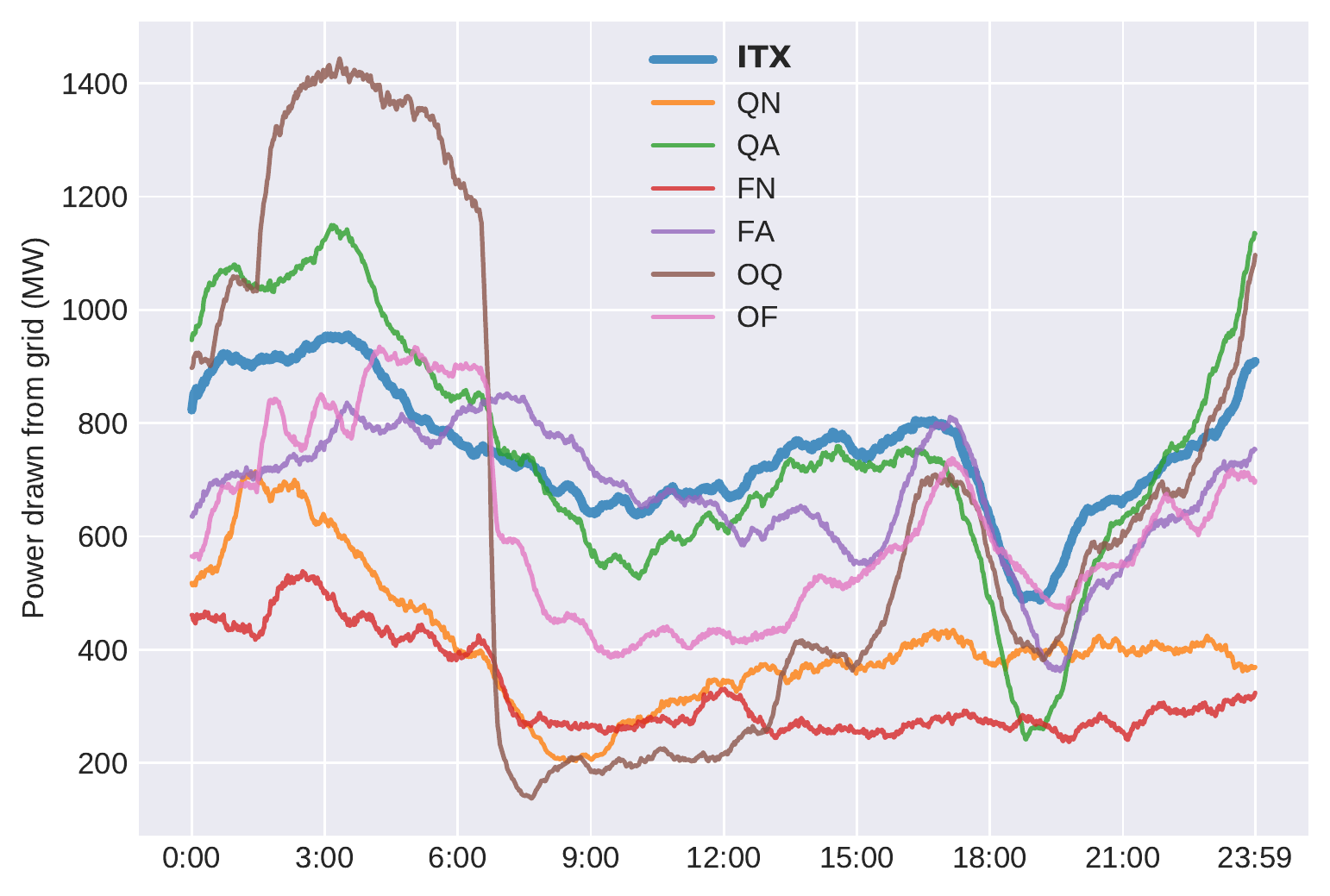}
    \caption{Grid load}
    \label{fig:gridpower}
  \end{subfigure}
  \caption{Performance comparison of charging strategies, aggregated by day of week (a, b, c) and by time of day (d, e, f)}
  \label{fig:resultsweek}
\end{figure*}

First, we look at the results of the simulator runs of the week of November 2-8. These simulator runs were executed with fleet size $|V| = 6000$. Figure \ref{fig:delay}, \ref{fig:ontimerate} and \ref{fig:customerspervehicle} show aggregated metrics per weekday, while Figure \ref{fig:stateofcharge}, \ref{fig:chargeroccupancy} and \ref{fig:gridpower} show the aggregated metrics for the time of day, i.e. from 0:00 until 23:59. Table \ref{tab:results} provides the comprehensive metrics for the complete simulator runs. The table also contains the increase and decrease percentages, which indicate the gains or losses that our proposed method yields in comparison to the baseline algorithms that were explained in Section \ref{sec:baselines}.

\subsubsection{Profitability}
The results in Table \ref{tab:results} show that our proposed method produces the highest monetary reward across the entire one-week period, i.e. a total amount of \$1,487,347. Hence, the ITX charging strategy is shown to outperform all the baseline strategies on the main objective function that we defined in Section \ref{sec:rewardfunction}. Compared to the most well-performing baseline, i.e. the Quick Charge, Available Station (QA) strategy, an increase in reward of nearly 5\% is observed, which amounts to a weekly advantage of \$70,000. This increase can be explained by the more optimal assignment that the ITX strategy performs in comparison to the QA strategy, as it produces a distribution in charging behaviour that better corresponds to the real-time demand patterns. This ensures that the charging sessions are spread out over time and space, and that future demand is anticipated when performing the charging control decisions. Consequently, this increases the overall fleet availability over time and facilitates more trips being served on-time, hence increasing the cumulative reward in the long term.

\subsubsection{Delays and on-time rate}
When we consider transport efficiency, the ITX strategy outperforms the baseline methods for two of the three metrics that were evaluated. The delay across the entire one-week simulation period was on average 0.805 minutes, which equals approximately 48 seconds. Even though it is a relatively small difference, the QA strategy produces a slightly higher average delay of 0.886 minutes, which is 5 seconds more than the ITX strategy produces. The improvement is more evident when looking at the on-time rate: with the ITX strategy, 95.78\% of the customers arrives with a delay of less than 5 minutes while the QN strategy yields an on-time rate of 89.30\%, which is 6.5\% lower. This suggests that a considerable improvement can be made when implementing ITX over the best-performing baseline, delivering 6.5\% more customers to their destinations within the acceptable delay. The difference between the on-time rate and the delay metric could be explained by the fact that the QN strategy produces a higher deviation of delays, resulting in a relatively low average delay while more customers are delivered with a delay above the five minute threshold. Meanwhile, for the ITX strategy, delays are observed to be more consistent. In comparison to the other baselines, large improvements are measured. For the overnight charging strategies OQ and OF, the on-time rate is 14.9\% and 20.1\% lower, respectively while the delays are 5 to 6 times higher than those of the ITX strategy. The accumulation of delays for the overnight charging strategies, especially the OF strategy, can be explained by the fact that it might not be possible to charge all vehicles with the necessary energy during the hours between 1:30 and 6:30. Some vehicles will therefore run lower on energy during the day and will have to perform an emergency charging session during the day. Additionally, during the weekend, demand in the night might be relatively high, producing higher waiting times when a significant part of the fleet is unavailable due to charging. The ITX, QA and FA strategies seem to handle these scenarios better as charging sessions are spread out more evenly over time, increasing the operational efficiency. Similar patterns are visible when looking at the delays and on-time rates per day of the week in Figure \ref{fig:delay} and \ref{fig:ontimerate}. For every day of the week, it is visible that the ITX and QA strategies consistently produce lower average delays than the other baselines, while the ITX strategy outperforms the QA strategy and all other baselines regarding on-time rate. Also, it emerges that the QN and FN strategies yield particularly high delays which peak on Thursday. This phenomenon is not visible for the other baselines, suggesting that the queues for some charging stations accumulate during the week and could not be elimated during times when demand is low, therefore gradually increasing the average delay to very high levels (between 30 and 35 minutes) in the second half of the week. The overnight strategies seem to produce peak delays during the Saturday, which is a logical consequence of the fact that demands during the night are larger on this day of the week.

\subsubsection{Passenger comfort}
Regarding passenger comfort (measured by the number of customers per vehicle), the ITX strategy performs slightly less well than the best-performing baseline QA. While ITX yields on average 0.611 customers per vehicle, QA performs slightly better with 0.607 customers per vehicle. However, it should be noted that the difference is very small, which is also visible in Figure \ref{fig:customerspervehicle}. From this figure, it emerges the average number of customers for ITX and QA is almost identical throughout the week. However, both strategies perform considerably better than the remaining baselines. The FA strategy results in 16.5\% more customers being carried in the same vehicle. The QN and FN baselines result in 95.6\% and 116.3\% more customers carried per vehicle, i.e. essentially a doubling of the number of customers in a single vehicle. This is likely caused by the fact that vehicles are unavailable for longer times due to long queues at charging stations, which drastically constrains the effective fleet size and forces more trips to be consolidated. Even though we allow the consolidation of trips in a ridepooling service, it remains essential to reduce the number of customers per vehicle, as this suggests that customers are more efficiently spread across the available fleet which increases the comfort experienced by customers of the service. The overnight charging strategies OQ and OF also yield a considerable higher number of customers per vehicle, i.e. 0.769 and 0.905 respectively. As visible in Figure \ref{fig:customerspervehicle}, this increase is most evident on Friday and Saturday, which is when higher demands during the night are observed. Since a considerable part of the fleet is charging overnight for the OQ and OF strategies, the remaining vehicles will need to serve the remaining trips and will therefore carry a higher number of customers on average.

\subsubsection{Energy consumption}
With regards to resource efficiency, we first look at the energy consumed per on-time request (in kWh) during the entire simulation period. In total, for the ITX strategy, 0.975 kWh is consumed on average per customer that was delivered on-time. Ideally, one would strive for the charging strategy that yields the lowest energy consumption, as this suggests that more efficient assignments (in terms of travel and queuing times) are produced. From Table \ref{tab:results}, it emerges that all baselines result in higher energy consumption per on-time request. For instance, with the FN strategy only 0.975 kWh per on-time request is consumed. It is $5.128\%$ higher for the best-performing baseline QA, and more than $20\%$ higher for the QN and FN baselines. This suggests that the ITX strategy is able to produce more optimal vehicle-charger assignments than the baseline strategies, reducing the necessary travel and queuing times and hence enabling lower energy consumption. Moreover, the PECT equation takes into account the impact of spatio-temporal demand fluctuations on the expected idle time after charging, aiming to reduce the distance to future demand areas after completing a charging session. This also enables a reduction of driving distance and energy consumption on the long term.

\begin{figure*}[t]
  \centering
  \includegraphics[width=0.88\linewidth]{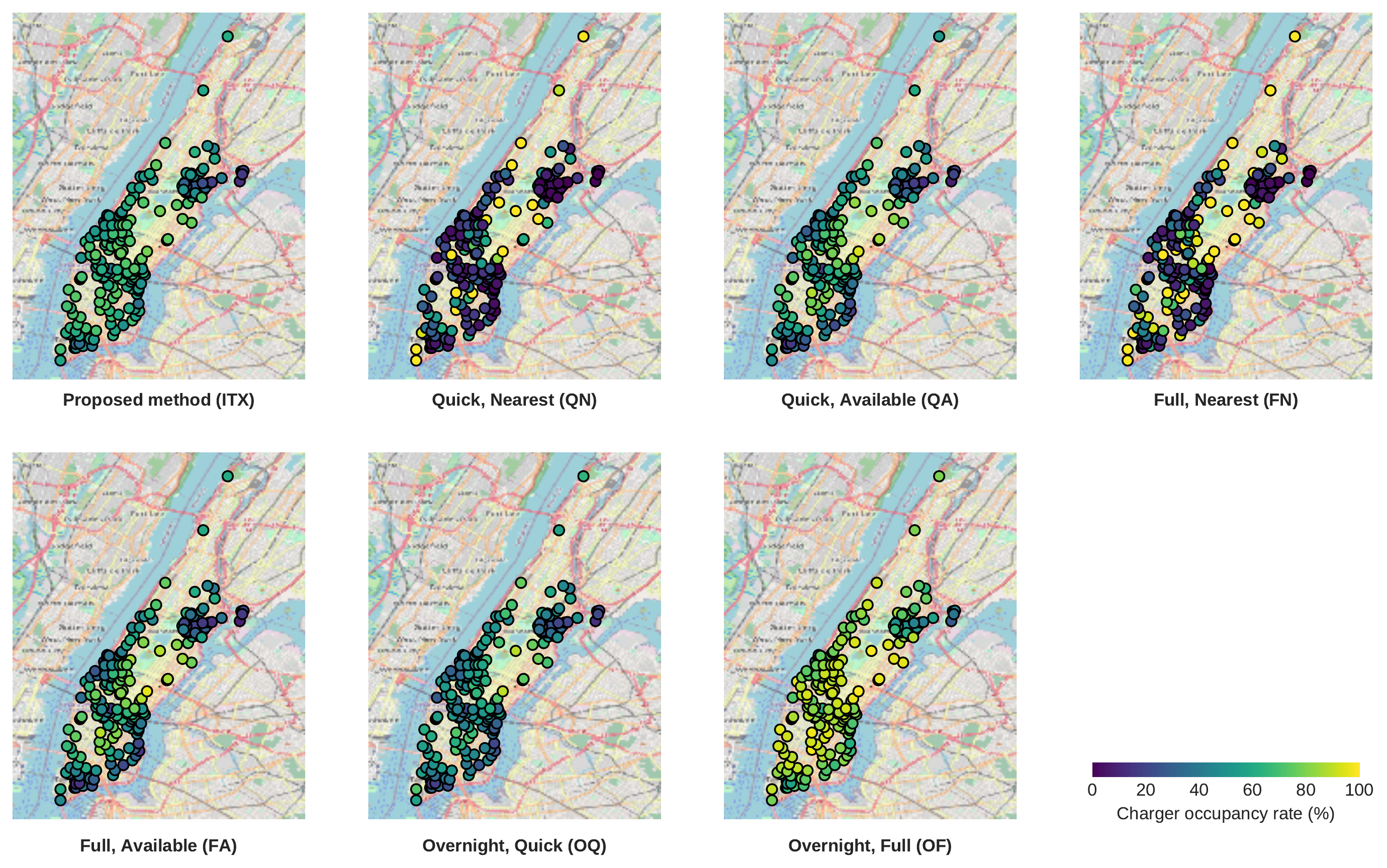}
  \caption{Comparison of charging station occupancy rates, averaged over the entire simulation period where $|S| = 500$}
  \label{fig:chargeroccmap}
\end{figure*}

\subsubsection{Battery levels and peak loads}
Looking at the State-of-Charge (SoC) over time as depicted in Figure \ref{fig:stateofcharge}, it is visible that the ITX strategy results in more stable energy levels than the baseline methods. In comparison to the QN and FN baselines, the average SoC is higher (on average 42\% instead of 24\% and 21\%), providing a higher reserve energy level for the vehicles. Merging this information with knowledge from other metrics, we argue that the SoC for the QN and FN strategies is so low due to the fact that vehicles spend long times in queues before being able to charge, which drops the average SoC for the fleet considerably. For the other baseline strategies, we see a clear increase in SoC just before evening rush hour at around 15:00, while the highest energy consumption is observed during rush hours (i.e. sharply decreasing SoC at around 7:00 and 18:00).  In comparison to the ITX strategy, we observe that the SoC levels for the QA strategy are located around the same average but show more fluctuations. The same phenomenon is generally visible with the FA and OF strategies, albeit with a higher SoC on average. This can be logically explained by the fact that these strategies perform a full charge (to 99\% SoC) instead of a `quick' charge (to 70\%), leading to larger energy reserves. However, the transport efficiency metrics suggest that the Full Charge baselines tend to produce higher delays due to their relatively long charging times. With the ITX strategy, the SoC levels are thus slightly lower but more uniform over time, suggesting that the idle times can be effectively used to keep the vehicles equipped with sufficient energy during the day. This finding also implies that the ITX strategy induces lower peak loads on the charging infrastructure, which is backed up by Figures \ref{fig:chargeroccupancy} and \ref{fig:gridpower}. From these figures, it becomes evident that the highest peak loads on the electricity grid are generated with the OQ baseline (1,450 MW), which is a logical consequence of the policy that all vehicles are charged overnight. With the ITX strategy, it emerges that this peak load can be reduced by 34.48\% to 950 MW. Compared to the generally best-performing baseline QA, peak load (1,150 MW) can be reduced by 17.39\%. It is visible that the FA and OF baselines both result in high occupancy rates but in relatively low peak loads, which seems counter-intuitive at first. This phenomenon can however be explained by the fact that the supplied power tapers off when the SoC gets above 70\%. As a result, peak power drawn from the grid will be lower if vehicles are fully charged instead of charged up to 70\% SoC. An incidental finding of this research is therefore that, even though quick charging generally improves transport efficiency, full charging sessions help to spread the load on charging infrastructure over time.

In Figure \ref{fig:chargeroccupancy}, a spatial overview of the charging station occupancy rates is provided. This allows us to observe how well the average charging load is spread out over space. Ideally, the load is equally distributed, as this reduces the burden on grid infrastructure in specific areas. The figure suggests that the ITX method yields more uniform distribution of charger occupancy over space. Especially for the QN and FN baselines, the variance in occupancy rates is very high, which can be explained by the fact that vehicles egoistically choose the nearest charging station, without taking the queue length or future demands into consideration. With the OQ, QA and FA strategies, it is evident that occupancy rates are slightly more uniform in space, arguably because the availability of charging stations is taken into consideration. The OF and ITX strategies both show even higher uniformity in space. The ITX strategy, however, produces a lower average occupancy rate than the OF strategy. This can be attributed to the fact that OF lets vehicles charge for a longer period, reducing their charging power and therefore resulting in more persistent charging occupancy rates. Overall, the ITX strategy results in the least grid load fluctuations (both temporally and spatially) in direct comparison to the other well-performing baselines QA, FA, OQ and OF. 

Combining this knowledge with other metrics, we argue that the ITX, QA and FA strategies perform best all-round and yield an optimal balance between transport efficiency and resource efficiency, which is supported by the overall reward shown in Figure \ref{tab:results}.

\subsubsection{Influence of charging infrastructure}
\begin{figure}[h]
  \centering
  \includegraphics[width=\linewidth]{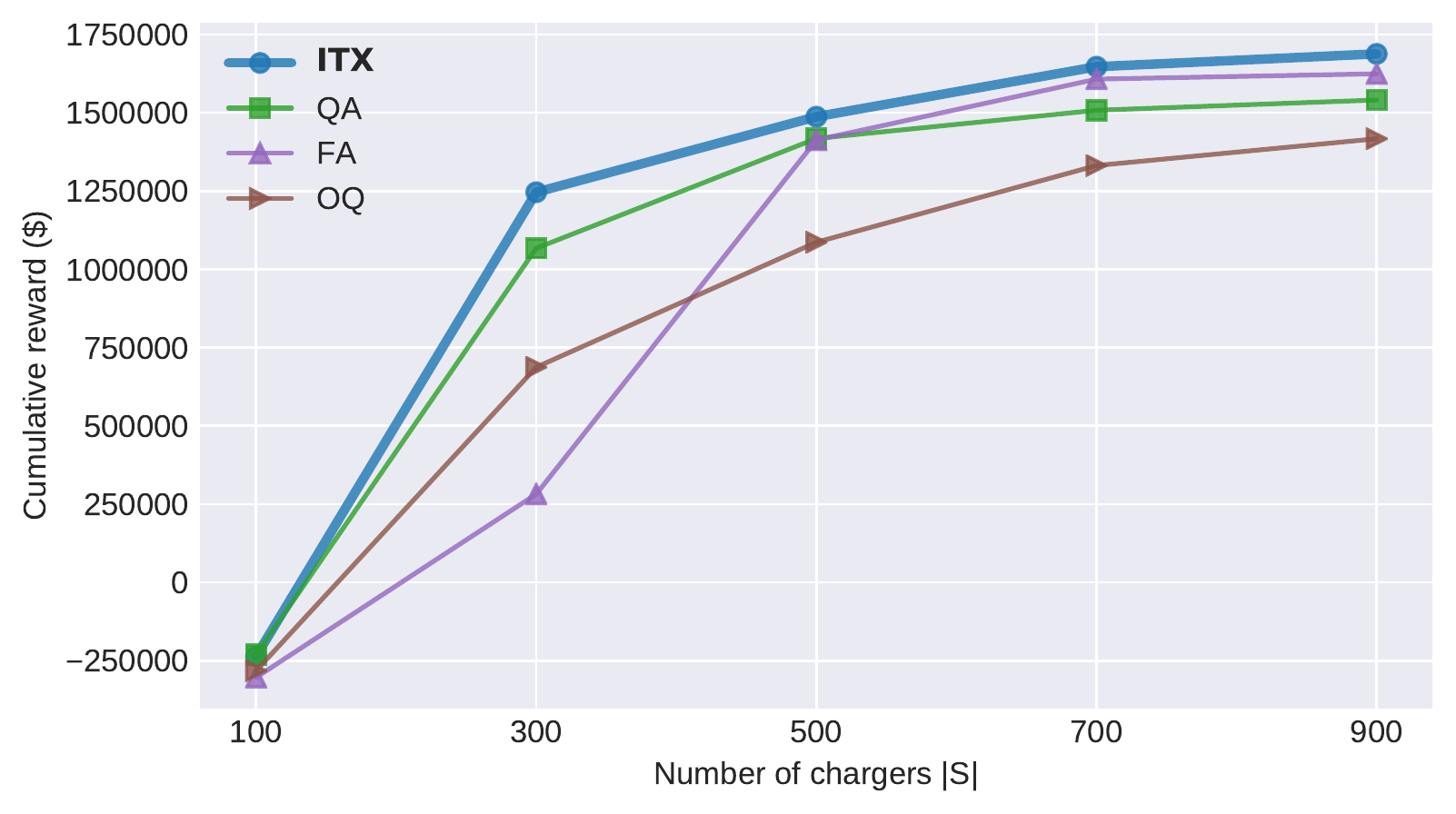}
\caption{Cumulative reward comparison for varying charging infrastructure}
\label{fig:rewardfleetsize}
\end{figure}

The results of the experiment with increasing number of chargers $|S|$ is displayed in Figure \ref{fig:rewardfleetsize}. Due to computational constraints, only the four best-performing strategies with regard to the overall objective function (i.e. ITX, QA, FA and OQ) were considered for this experiment. The figure demonstrate that, for all strategies considered, the reward increases rapidly between 100 and 500 chargers, after which the reward flattens out. For a low number of chargers, i.e. where $|S| < 500$, it is visible that ITX and QA outperform the FA and OQ baselines by a considerable margin, with ITX being the leading strategy across the entire experiment. The margins become thinner as the number of chargers grows, which can be explained by the fact that more charging infrastructure will lead to lower queuing times. Therefore, even with a suboptimal charging strategy, the magnitude of the charging infrastructure will compensate for relatively inefficient decision-making. The results demonstrate that our proposed method ITX can produce a weekly reward of $\$1,246,088$ for a relatively low number of chargers $|S| = 300$, which is approximately similar to what the OQ strategy achieves for a much higher number of chargers $|S| = 700$. From an operational perspective, this suggests that, when replacing the OQ strategy by the ITX strategy, more than half of the charging infrastructure can be removed while achieving similar levels of profitability. This could arguably facilitate advancements regarding (cost-)efficiency and sustainability. Additionally, for $|S| = 300$ the results demonstrate that a $16.75\%$ increase in profitability can be achieved in comparison to the QA baseline, amounting to $\$178,907$ on a weekly basis with a fleet of 6,000 vehicles. All things considered, we argue that our proposed method is especially suitable for extracting the maximum performance out of limited infrastructure.

\subsubsection{Runtime tests}
\begin{figure}[h]
  \centering
  \includegraphics[width=\linewidth]{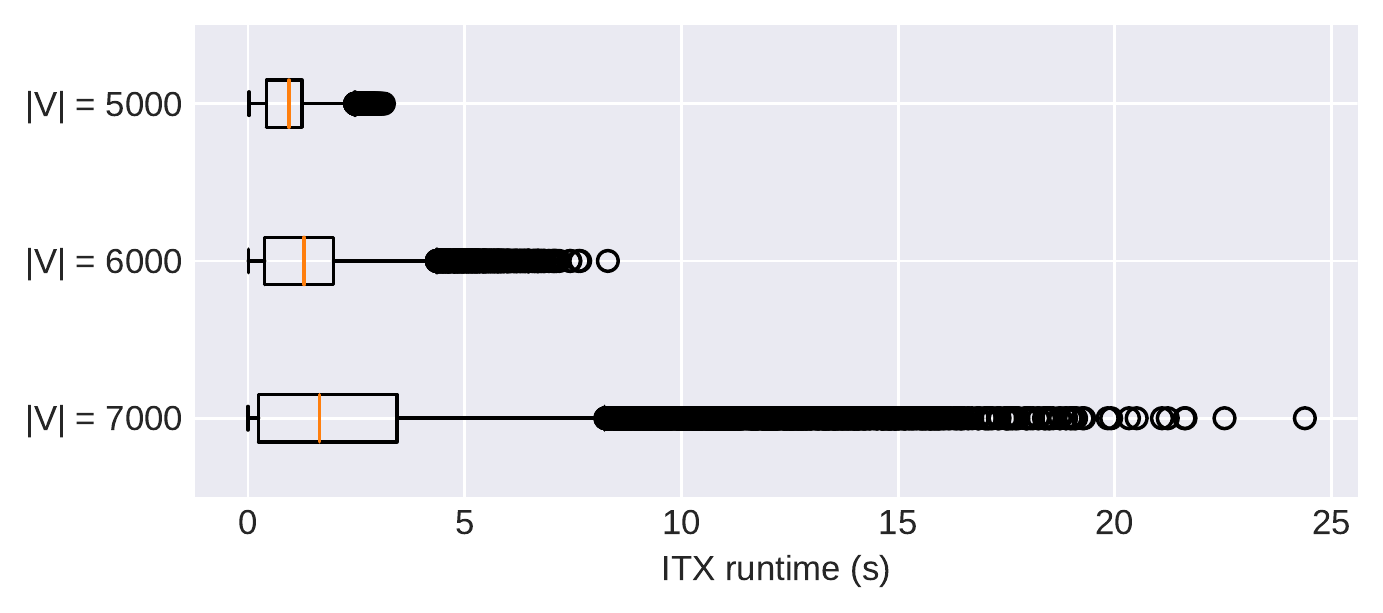}
\caption{Runtimes of ITX algorithm with varying fleet size}
\label{fig:runtime}
\end{figure}
The results of the runtime tests with increasing fleet sizes is displayed in Figure \ref{fig:runtime}. With fleet size $|V| = 5000$, the mean runtime is $0.967$ seconds (with a standard deviation of $0.658$), while the peak runtime reaches up to $3.135$ seconds. When increasing the the fleet size to $|V| = 6000$, the mean runtime increases to $1.573$ seconds (with a standard deviation of $1.476$) while the highest outlier is located at $8.302$ seconds. Further increasing the fleet size to $|V| = 6000$ increases the mean runtime to $2.970$ seconds (with a standard deviation of $3.785$) and the maximum runtime to $24.384$. By extrapolation, we therefore devise that the algorithm runs in polynomial time and suffices the real-time constraint of 60 seconds when fleet size $|V| < 6500$. A possibility to further reduce computational times is to use heuristics to compute the assignment instead of the Hungarian algorithm.

\section{Conclusion \& Future Work}
\label{sec:conclusion}
In this work, we have presented a predictive charging method for ridepooling services. This method relies on Graph Convolutional Networks (GCNs) and a linear assignment algorithm to devise an optimal pairing of vehicles and charging stations. We first predict the (remaining) idle times for all idle vehicles using a GCN, after which we compute the potential effective charging time (PECT) for every possible pairing and solve the resulting assignment problem iteratively. The main idea behind this approach is that predicted idle times can be proactively exploited to charge, such that unprofitable periods are effectively used to harvest energy and charging sessions can be spread more evenly across time and space. Furthermore, the method enables the combination of three different decision variables (when, where and for how long to charge) into a comprehensive optimization task that is solvable in polynomial time. Our results show that the approach is capable of running in real-time on large-scale networks of 6,000 vehicles and 8,500 roads, representing the complete road network of Manhattan, New York City. We demonstrate using a real-world trip request dataset that our proposed method is able to deliver fine-grained and precise decision-making at the real-world scale level of a ridepooling service. 

The proposed Idle Time Exploitation (ITX) strategy outperforms all baselines in terms of the overall reward (expressed in US Dollars). Compared to the best performing baseline strategy, where vehicles are charged up to 70\% SoC to the nearest available station, an improvement of nearly 5\% can be gained, which amounts to a weekly gain in operational profits of approximately \$70,000 for an operation of 6,000 vehicles. In comparison to the overnight charging strategy, which is prevalent in many electric transportation systems, the ITX strategy can provide a weekly increase of 36.9\% in profitability, amounting to approximately \$400,000 on a weekly basis. When looking closer at transport efficiency metrics, it emerges that the usage of the ITX strategy does not significantly reduce the average delays compared to the best baseline strategy. However, the percentage of trips that is completed with a delay lower than 5 minutes (i.e. the on-time rate) is increased by at least 6.5\% when using ITX for charging control. Furthermore, customer comfort levels are higher with ITX than with any of the baseline strategies, facilitating a better spread of customers across the fleet such that vehicle occupancy remains as low as possible. These metrics suggest that ITX positions vehicles better after charging, enabling them to anticipate and act upon new demand areas after finishing a charging session.

Regarding resource efficiency, the results suggest that the ITX strategy is able to conserve more stable energy levels throughout the fleet, therefore providing better reserves during periods where demand is particularly high. This is further backed up by the observation that charger occupancy rates and grid loads are more uniform over time.  For instance, the peak load on the electricity grid is approximately 950 MW for the ITX strategy, while it is 1,450 MW for the OQ strategy and 1,150 MW for the QA strategy. Lower peak loads alleviate the burden on the electrical grid infrastructure, providing benefits to the grid operator while also yielding better resource efficiency for the operator of the ridepooling service. After all, stable energy reserves help to absorb high peaks in demand, facilitating better availability of the fleet and hence improvements of on-time performance and customer comfort.

In the future, we aim to study how the idle time exploitation technique can be used to combine the dispatch, repositioning and charging processes into a comprehensive optimization task. We argue that the three processes, although modeled separately, are highly intertwined and that the outcomes of one process influence the other two. For instance, we believe that there are opportunities to use the prediction model to quickly evaluate where vehicles could reposition to reduce the expected idle times. Also, upon dispatch, the expected idle time after dropoff of all customers may influence the optimal strategy, essentially integrating the repositioning process into conventional dispatch control. We therefore argue that the ITX methodology might be further developed for simultaneous control of dispatching, repositioning and charging strategies. This could provide operators and service providers with decision-making capabilities that yield close-to-optimal results while allowing for modular integration into the existing operations. We also aim to study the transferability of the ITX solution to other regions and cities, as we argue that the real-time and data-driven nature of this solution facilitates easy application of our proposed methodology in different scenarios. Another research gap arises concerning the implementation of variable energy costs, which is a contemporary measure taken by grid operators to reduce peak loads on electricity infrastructure. One could evaluate the resilience of ITX to these dynamic pricing models in comparison to the baseline strategies. Furthermore, we aim to study how the ITX strategy reacts to days with special demand patterns, such as holidays or large events. Since the algorithm is driven by a neural network that predicts idle times, it is likely that these predictions will become less accurate on such occasions. We argue that the proposed method is better at building up energy reserves in the fleet, making it possible to handle unexpected surges in demand. However, it is still of interest to assess empirically whether ITX is able to handle unconventional demand patterns better than baseline models. Additionally, we aim to research whether ITX is able to handle a higher diversity of vehicles in the fleet, with differences in battery capacities and charging power. We are interested whether ITX is able to structurally outperform the baseline models in scenarios where battery capacities are highly heterogeneous and charging power is limited. Lastly, a possibly interesting research direction could be to evaluate the impact of renewable energy production on the performance of the ITX charging method. For instance, it could be interesting to simulate the generation of solar energy at charging stations and assess whether the use of ITX leads to more efficient use of generated energy during peak sunlight hours, hence mitigating the need for expensive high-capacity energy storage. This way, our proposed methodology could contribute to the improvement of efficiency for transport, infrastructure and grid operators, reducing the operating cost of ridepooling systems while alleviating burdens on road and grid infrastructure and hence paving the way for more sustainable transport systems.

\bibliographystyle{IEEEtran}
\bibliography{sample-bibliography.bib}

\newpage
 
\vspace{11pt}

\begin{IEEEbiography}[{\includegraphics[width=1in,height=1.25in,clip,keepaspectratio]{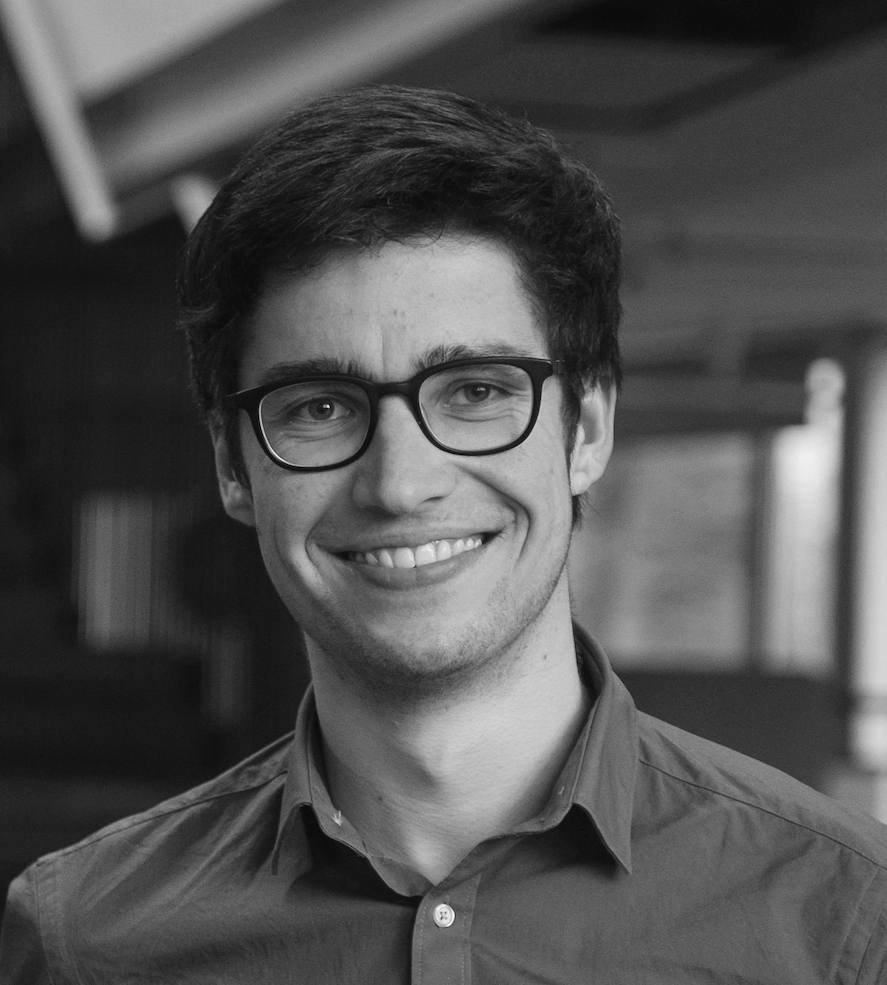}}]{Jesper C. Provoost}
is currently working as a PhD researcher in the Department of Transport \& Planning at Delft University of Technology. He received his Master's degree in Computer Science from the University of Twente in 2021. His research interests include operations research, simulation-based optimization, reinforcement learning and predictive modeling.
\end{IEEEbiography}

\vfill

\end{document}